\documentclass{clv2025}

\usepackage{amsmath}
\usepackage{booktabs}

\usepackage{enumitem}
\usepackage{graphicx}
\usepackage{booktabs}    
\usepackage{subcaption}  

\usepackage[most]{tcolorbox} 

\usepackage{xcolor}
\definecolor{cmint}{RGB}{0,128,128}      
\definecolor{corange}{RGB}{255,140,0}   
\definecolor{textbox}{RGB}{220,220,220}  
\definecolor{headerbox}{RGB}{128,128,128}

\usepackage{lipsum} 

\usepackage{booktabs}   
\usepackage{multirow}   
\usepackage{makecell}   
\usepackage{adjustbox}  
\usepackage{pifont}
\usepackage{hyperref}
\usepackage{float}

\newcommand{\AppendixSection}[2][]{%
  \renewcommand{\thesection}{\Alph{section}}
  \refstepcounter{section}
  \phantomsection
  \section*{Appendix \thesection: #2}%
  \ifx\relax#1\relax
    \addcontentsline{toc}{section}{Appendix \thesection: #2}%
  \else
    \addcontentsline{toc}{section}{#1}%
  \fi
}

\usepackage{fancyvrb}

\tcbuselibrary{listings,skins,breakable}

\usepackage[utf8]{inputenc}
\usepackage{listings}
\tcbuselibrary{listings,skins,breakable}

\lstdefinestyle{jsonwrap}{
  basicstyle=\ttfamily\small,
  breaklines=true,
  breakatwhitespace=true,
  columns=fullflexible,
  showstringspaces=false
}

\runningtitle{Beyond Binary Classification}
\runningauthor{De Grazia et al.}

\begin{document}

\title{Beyond Binary Classification: Detecting Fine-Grained Sexism in Social Media Videos}


\author{Laura De Grazia$^{1}$, 
Danae Sánchez Villegas$^{2}$, 
Desmond Elliott$^{2}$, 
Mireia Farrús$^{1}$, 
Mariona Taulé$^{1}$}


\affilblock{
    \affil{CLiC – Language and Computing Center, University of Barcelona\\\quad \email{lauradegrazia@ub.edu}, \email{mfarrus@ub.edu}, 
    \email{mtaule@ub.edu}}
    \affil{Department of Computer Science, University of Copenhagen\\\quad \email{davi@di.ku.dk}, \email{de@di.ku.dk}}
}

\maketitle

\begin{abstract}

Online sexism appears in various forms, which makes its detection challenging. Although automated tools can enhance the identification of sexist content, they are often restricted to binary classification. Consequently, more subtle manifestations of sexism may remain undetected due to the lack of fine-grained, context-sensitive labels. To address this issue, we make the following contributions: (1) we present FineMuSe, a new multimodal sexism detection dataset in Spanish that includes both binary and fine-grained annotations; (2) we introduce a comprehensive hierarchical taxonomy that encompasses forms of sexism, non-sexism, and rhetorical devices of irony and humor; and (3) we evaluate a wide range of LLMs for both binary and fine-grained sexism detection. Our findings indicate that multimodal LLMs perform competitively with human annotators in identifying nuanced forms of sexism; however, they struggle to capture co-occurring sexist types when these are conveyed through visual cues.\footnote{The dataset will be made available at \url{https://github.com/lauradegrazia/FineMuSe}.}
\textcolor{red}{\textbf{Warning}: \textit{This paper contains examples of language and images which may be offensive}.}


\end{abstract}

\section{Introduction}\label{sec1}
Sexism is generally understood to be prejudice and discrimination based on sex or gender \cite{samory2021call}.
It can take several forms, appearing within social institutions, for example, in the phenomenon of the gender pay gap \cite{alkadry2006unequal}; in social interactions, where it is normalized and reproduced through everyday behaviors; and at the individual level, when people express sexist beliefs as part of their self-identity construction \cite{homan2019structural}. 

Extensive research has focused on analyzing sexism toward women, but sexism affects individuals of all genders, including non-binary people and those within the LGBTQ+ community \cite{o2009encyclopedia, javidan2021structural, luo2025literature}. In this work, we consider the impact of sexism toward women and Gender Non-Conforming identities, i.e., individuals who do not conform to the cisgender model \cite{grazia2025mused}. Social media platforms amplify the impact of sexism in multiple ways, such as the promotion of toxic masculine behaviors \cite{papadamou2021over}, the rejection of feminism \cite{samory2021call}, or the reinforcement of body standards to which women are expected to conform \cite{sarda2025social}. Automated tools for recognizing sexism are widely used to support the work of content moderators; however, these tools are often limited to binary classifications \cite{dev2025beyond} or provide only high-level categories such as toxicity, abuse, or sexism \cite{kirk2023semeval}. As a result, more subtle and implicit forms of sexism may go unrecognized due to the absence of fine-grained and context-sensitive labels. Furthermore, automated tools that flag content as sexist without providing a clear explanation lack transparency \cite{samani2025large}, diminishing trust in automated moderation tools among both users and moderators. Previous studies that adopted a fine-grained approach to sexism detection have mainly focused on textual content, primarily using data from X \cite{anzovino2018automatic, rodriguez2021overview} and Reddit \cite{zeinert2021annotating, guest2021expert}. Current research applying granular labels to multimodal sexism detection has mostly concentrated on memes, analyzing the relationship between text and image \cite{plaza2024exist, singh2024mimic}. In contrast, the detection of fine-grained sexism in videos, considering the interaction between text, speech and image, is an under-explored area of research. Recent work on sexism detection in social media videos has primarily relied on binary classification, with \citet{grazia2025mused} showing that extended definitions of sexism improve detection performance for both human annotators and LLMs compared to a simple definition. Specifically, performance improves when the definition captures the various levels of granularity through which sexism can manifest, including forms such as stereotyping or the denial of gender inequality. Although the study by \citet{arcos2024sexism} incorporates fine-grained labels based on the theoretical framework developed by \citet{plaza2024exist}, it relies on data from only one social media platform (TikTok). In abusive language detection it is important to sample from multiple platforms, as different social media attract distinct user communities and each platform constitutes its own linguistic domain \cite{karan-snajder-2018-cross, zeinert2021annotating}. Finally, recent research on the evaluation of generated explanations by automated systems has focused primarily on hate speech detection \cite{wang2023evaluating}, whereas the analysis of explanations in sexism detection has not been explicitly explored. 

Building on these insights, we introduce FineMuSe, a new fine-grained multimodal dataset for sexism detection in Spanish. FineMuSe builds upon the foundations of MuSeD, which focused on binary sexism classification using TikTok and BitChute videos \cite{grazia2025mused}. Our dataset extends MuSeD along several key dimensions. First, we increase platform diversity by adding 428 additional videos collected from YouTube Shorts, expanding the representation of short-form social media content. Second, we annotate all content with fine-grained labels that capture distinct forms of sexism, including stereotypes, inequality, discrimination, and objectification, as well as non-sexist content and the use of rhetorical devices (irony and humor), as detailed in Section~\ref{sec:taxonomy}. The fine-grained annotation was conducted across multiple modalities (text, audio, and video), extending the MuSeD annotation scheme, which applies this multi-level approach only to the binary classification of content as sexist or non-sexist. Third, we relied exclusively on expert annotators, who were highly trained and had prior experience in annotating sexist content. These factors contributed to the creation of a high-quality dataset.
Our main contributions are as follows:

\begin{itemize}

\item We present FineMuSe, a multimodal sexism detection dataset in Spanish with binary and fine-grained annotations. FineMuSe is a multi-source dataset that includes 828 videos collected from TikTok, BitChute, and YouTube Shorts. 
\item We develop a detailed hierarchical taxonomy for labeling sexism online grounded on previous studies and examining the collected data.
\item We provide an in-depth analysis of the dataset characteristics and the ways sexism manifests across different modalities and platforms. Moreover, we examine the distribution of regional varieties of Spanish in FineMuSe. 
\item  We have conducted multimodal experiments to benchmark performance across modalities and levels of granularity. 
\item We perform a correlation analysis to assess how the predictions regarding sexist types produced by the best-performing fine-grained detection model correspond to human annotations.
\item Finally, we examine model justifications and human preferences, analyzing whether humans prefer model or label-based rationales when evaluating instances of sexism and non-sexism. 


\end{itemize}

\section{Background and Related Work}\label{sec2}  

\subsection{Resources for Identifying Sexist Content}\label{subsec2}
\paragraph{Text and Image Resources} 
\citet{gasparini2018multimodal} introduce SAD: the Sexist Advertisement Database. This database includes a dataset with 423 advertisements labeled as sexist or non-sexist based solely on their visual content; and a second dataset that contains 192 advertisements labeled as sexist or non-sexist according to visual and/or textual cues. \citet{fersini2019detecting} present the MEME dataset, which comprises 800 memes labeled as sexist or non-sexist, with the sexist memes further annotated for aggressiveness and irony. The SemEval-2022 Task 5: Multimedia Automatic Misogyny Identification (MAMI) \cite{fersini-etal-2022-semeval} includes a benchmark dataset of 15,000 memes annotated as misogynistic or non-misogynistic, and further categorized into four types of misogyny (Shaming, Stereotype, Objectification, and Violence). This study highlights that the fine-grained classification task is more challenging than the binary task, as reflected in the inter-annotator agreement (IAA), which ranges from moderate for the binary task (k = 0.58) to fair for the fine-grained task (k = 0.34).
\citet{buie2023social} introduce the Social Media Sexist Content (SMSC) Database, a resource comprising memes, personal posts, and graphics designed to support the identification of sexist content. To evaluate the material, the study draws on Ambivalent Sexism \cite{glick1997hostile} and Objectification theories \cite{fredrickson1997objectification}. Finally, the EXIST dataset 2024 comprises 2,000 memes in English and Spanish for the training set and 500 memes per language for the test set \cite{plaza2024exist}.


\paragraph{Video Resources} The identification of sexism in videos using a fine-grained approach is still in its early stages. Most previous research has focused on hate speech \cite{das2023hatemm, wang2024multihateclip, wang2025hateclipseg}, which includes hateful content related to sex and gender. However, sexism can manifest in more implicit and nuanced forms compared to hate speech. Current research on detecting sexism in videos includes the work of \citet{arcos2024sexism} and \citet{grazia2025mused}. \citet{arcos2024sexism} introduce a multimodal dataset focused on detecting discriminatory content against women. This dataset comprises \textasciitilde13 hours of Spanish videos and \textasciitilde11 hours of English videos, fine-grained annotated using the framework developed by \citet{plaza2024exist}. 

\citet{grazia2025mused} introduce MuSeD (Multimodal Dataset for Sexism Detection), which comprises \textasciitilde11 hours of videos collected from TikTok and BitChute. This study adopts a comprehensive definition of sexism, addressing issues related to sex, sexual orientation and gender identity. MuSeD features a multi-level annotation method for labeling sexist content across various modalities, including text, audio and video. 

\paragraph{Limitations of Existing Resources}
Table \ref{tab:table_dataset_comparison} compares existing multimodal datasets for analysing sexism and misogyny. We observe that most current work focuses on fine-grained misogyny detection in memes. In the field of sexism detection in videos, we found that \citet{arcos2024sexism} used fine-grained categorizations, but they sampled only from one platform and restricted their definition of sexism to discrimination against women. In comparison to previous datasets, FineMuSe introduces the following novelties: (i) We examine diverse categorizations of both sexism and non-sexism, analyzing not only the forms through which sexism spreads but also how individuals respond to it, whether through overt dissent or by sharing personal experiences; (ii) We include annotations of irony and humor for both sexism and non-sexism, allowing us to analyze how sexist and non-sexist content use different rhetorical strategies; (iii) We expand the annotation framework introduced by \citet{grazia2025mused}, examining how annotators recognize fine-grained types of sexism in different modalities; (iv) We analyze the distribution of different regional varieties of Spanish.


\begin{table*}[t!]
\centering
\caption{A comparison of multimodal datasets for sexism and misogyny analysis. 
W = Women, FB = Facebook, IG = Instagram, R = Reddit, P = Pinterest, TT = TikTok, YT = YouTube, S = Sexist, NS = Non-Sexist, M = Misogyny, NM = Non-Misogynistic, H = Hostile Sexism, B = Benevolent Sexism, O = Objectification, A = Aggressiveness, I = Irony, HU = Humor. Numbers in brackets indicate the number of categories per label.}
\small
\setlength{\tabcolsep}{2pt} %
\renewcommand{\arraystretch}{1.2} %
\resizebox{\textwidth}{!}{%
\begin{tabular}{@{}lcccccc@{}}
\toprule
\textbf{Dataset} & \textbf{Platform} & \textbf{Language} & 
\makecell{\textbf{Target} \\ \textbf{group}} & 
\makecell{\textbf{Annotation} \\ \textbf{of modalities}} & 
\makecell{\textbf{Annotators} \\ \textbf{expertise}} & 
\textbf{Label} \\
\midrule
\multicolumn{7}{l}{\textbf{Advertisements}} \\
\midrule
\citet{gasparini2018multimodal}  & FB & eng & W & \ding{55} & Not Specified & S, NS \\
\midrule
\multicolumn{7}{l}{\textbf{Memes}} \\
\midrule
\citet{fersini2019detecting}  & FB, X, R, IG & eng & W & \ding{55} & Crowd Sourcing & S(A,I), NS \\
\citet{fersini-etal-2022-semeval}  & X, R, MGTOW & eng & W & \ding{55} & Crowd Sourcing & M(4), NM \\
\citet{buie2023social} & Pixabay & eng & W & \ding{55} & Crowd Sourcing & H, B, O \\
\citet{singh2024mimic} & FB, IG, R, P & hin, eng & W & \ding{55} & Trained annotators & M(3), NM \\
\citet{ponnusamy-etal-2024-laughter} & IG, FB, P & tam, mal & W &\ding{55} & Trained annotators & M, NM  \\
\citet{plaza2024exist} & Google Images & eng, spa & W & \ding{55} & Crowd Sourcing & S(5), NS \\
\midrule
\multicolumn{7}{l}{\textbf{Videos}} \\
\midrule
\citet{arcos2024sexism} & TT & eng, spa & W & \ding{55} & Trained annotators & S(5), NS \\
\citet{grazia2025mused} & TT, BitChute & spa & W, LGBTQ+ & \ding{51} & 
\makecell{One expert annotator,\\trained annotators} &  S, NS \\
Fine-MuSe (Ours) & TT, BitChute, YT  & spa & W, LGBTQ+ & \ding{51} & 
\makecell{Expert annotators} &  S(4), NS(2), I, HU \\
\bottomrule
\end{tabular}%
} 
\label{tab:table_dataset_comparison}
\end{table*}


\subsection{Fine-Grained Sexism Detection}\label{MM}

Previous studies have relied primarily on text-based models \cite{luo2025literature, samani2025large}, while the identification of the various manifestations of sexism in multimodal content remains largely unaddressed. Fine-grained prediction in multimodal data has mostly been explored within the realm of memes, where models struggle due to a lack of cultural understanding, for instance, failing to recognize instances of fat-shaming, or when text and image convey different interpretations \cite{fersini-etal-2022-semeval}. 

Research by \citet{grazia2025mused} indicates that multimodal models surpass text-only approaches, underscoring the need to incorporate multiple modalities for effective sexism detection. However, their study is limited by being based on a binary classification. Building on these insights, here we introduce FineMuSe, a multimodal dataset for sexism detection in Spanish with fine-grained labels. FineMuSe enables a more accurate evaluation of multimodal models tasked with predicting granular forms of sexism.


\subsection{Automatic Explanations for Sexism Detection}\label{Evaluation}

Automatic explanations for sexism detection aim to enhance the accuracy and transparency of automated systems by providing clear justifications for why specific content is flagged as sexist \cite{samani2025large}. 
\citet{samani2025large} demonstrate that Mistral-7B and LLaMA-3-8B outperform traditional models in classifying fine-grained forms of sexism. \citet{rayhana2025interpretable} employ LIME, a local explanation method, to make the predictions of the model (whether a given statement is sexist or non-sexist) more interpretable to humans. However, these efforts do not explicitly analyze the automatically generated explanations produced by the models themselves. Such analysis has only recently begun to emerge in the context of hate speech detection. \citet{wang2023evaluating} conduct a human study assessing the quality of the explanations generated by GPT-3 for hateful and non-hateful content. Their results show that GPT-3 is competitive at producing high-quality explanations, but that it must be used with caution in content moderation settings, as biased explanations may mislead human judgment. Inspired by these findings in hate speech detection, we undertake a detailed analysis of the quality of the explanations generated by models, comparing them with human-written explanations for sexist and non-sexist video content (see Section \ref{subsec:just_quality}).

\section{FineMuSe: a Fine-Grained Multimodal Dataset for Sexism Detection}\label{sec22}
We present FineMuSe, a multimodal dataset in Spanish for sexism detection, consisting of content annotated at a binary, fine-grained level across multiple modalities (text, speech, and video). 

\subsection{Data Sources}
\label{sec:data_sources}
FineMuSe is a multi-source dataset that extends the MuSeD dataset \cite{grazia2025mused} by adding 428 videos sampled from YouTube. MuSeD includes binary annotations (sexist and non-sexist) of 400 videos: 367 collected from TikTok, which explicitly bans misogyny and misgendering,\footnote{\url{https://www.tiktok.com/community-guidelines/en}} and 33 from BitChute, a video-hosting platform with a high prevalence of hateful content \cite{ das2023hatemm}. The YouTube subset consists exclusively of \textit{shorts}, a video format designed to closely resemble TikTok content. Unlike standard YouTube videos, which can range from a few seconds to several hours, \textit{shorts} are limited to a maximum duration of 60 seconds and must be produced in vertical or square format, making them particularly suited for mobile consumption. Since its introduction on YouTube, the format has gained popularity among content creators \citep{violot2024shorts}. While previous research has mainly examined hate speech and toxic content in traditional YouTube videos \cite{papadamou2021over, wang2024multihateclip, wang2025hateclipseg}, our focus on \textit{shorts} enables cross-platform analysis of two moderated short-form video platforms (TikTok and YouTube Shorts) and  a long-form platform with minimal content moderation (BitChute) \cite{trujillo2020bitchute, trujillo2022mela}.

\subsection{Data Collection}
\label{sec:data_collection}

YouTube Shorts were collected in April and May 2025. To retrieve the videos, we followed the sampling strategy introduced in \citet{grazia2025mused}. This approach consists of collecting video URLs using a set of Spanish hashtags.\footnote{The complete list includes 187 hashtags and is available in Appendix~\ref{app:hashtags}.} Since FineMuSe encompasses both Peninsular and Latin American Spanish, we conducted a detailed analysis of the regional varieties represented (see Section~\ref{subsec:Spanish_analysis}). After collecting the video URLs, the videos were downloaded using the Apify platform \citep{arcos2024sexism, grazia2025mused}, ensuring that only publicly available content was included. 
For the collection of \textit{shorts}, we focused on a set of themes aimed at capturing fine-grained and nuanced forms of sexism. These include the use of gender stereotypes (e.g., \#estereotipodegénero: gender stereotype), the denial of gender inequality and rejection of feminism (e.g., \#brechadegénero: gender gap; \#feminismotóxico: toxic feminism), discrimination based on sexual orientation and gender identity (e.g., \#ideologiadegénero: gender ideology; \#identidaddegénero: gender identity), and the promotion of toxic masculinity that portrays women as objects to be conquered (e.g., \#machoalpha: alpha male; e.g., \#alfaseductor: alpha male seducer). 

\subsection{Data Processing}
We transcribed the audio into text using Whisper-ctranslate2\footnote{\url{https://github.com/Softcatala/whisper-ctranslate2}} and extracted the audio from the videos with the \texttt{FFmpeg} software.\footnote{\url{https://www.ffmpeg.org}} The use of Whisper-ctranslate2, which generates transcripts including speaker diarization and timestamps, significantly improved the annotation process for both binary categorization and fine-grained annotation. In the binary annotation task, it enabled annotators to identify, within dialogues, which speaker produced sexist statements and which did not. In the fine-grained annotation task, it allowed annotators to accurately select the specific text segment, known as a ``span'', to which labels were applied.

\begin{figure}[t]
    \centering
    \caption{Three-level hierarchical taxonomy for labeling sexism, non-sexism, and rhetorical devices in social media videos.}\includegraphics[width=0.7\linewidth]{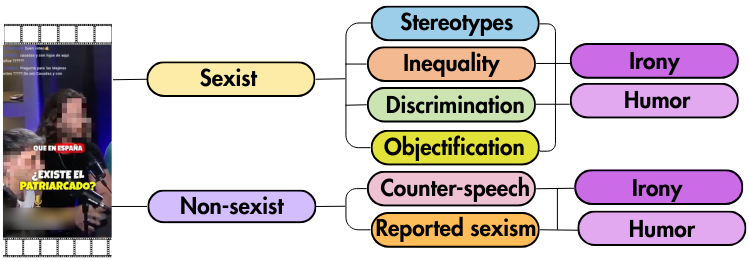}
    \label{fig:taxonomy}
\end{figure}

\section{Taxonomy and Annotation Process}
\label{sec:taxonomy}
Most previous work uses mutually exclusive categories \cite{jha2017does, anzovino2018automatic, mulki2021let}; in contrast, we adopt a multi-label approach, as different forms of sexism often overlap. The taxonomy was developed by combining a deductive approach, grounded in prior research and informed by reviews of existing studies on sexism, misogyny, discrimination based on gender identity and sexual orientation, and the annotation of toxic language. We also adopted an inductive approach, based on examining the collected data and engaging in discussions between authors and annotators \cite{zeinert2021annotating}. This dual approach allowed us to construct a robust taxonomy that reflects the current nature of the collected data and is replicable for future studies (see Figure~\ref{fig:taxonomy}). Our taxonomy has three hierarchical levels. First, we implemented a binary distinction between Sexist and Non-sexist content, which are mutually exclusive. Second, we performed a subclassification within both sexist and non-sexist content. For Sexist content, we defined four categories: (i) Stereotypes; (ii) Denial of Inequality and Rejection of Feminism; (iii) Discrimination based on sexual orientation, gender identity, and the LGBTQ+ community; and (iv) Objectification. For Non-sexist content, we defined two categories: (i) Counter-speech against sexism; and (ii) Report of sexist experiences. Third, we included a classification of the rhetorical devices of Irony and Humor, applicable to both sexist and non-sexist content. The second-level categories are not mutually exclusive, meaning that multiple labels can be assigned to the same content. For example, a sexist instance can be categorized simultaneously as both Stereotypes and Denial of Inequality and Rejection of Feminism. 

\begin{table*}[t]
    \caption{Examples of \textit{short} videos by sexism type along with their text transcriptions. Example (a) includes a stereotype, reinforcing the traditional gender division between men (external power) and women (domestic sphere). Example (b) expresses the denial of inequality, conveying the idea that Spanish law marginalizes men and overprotects women. Example (c) illustrates discrimination by claiming that trans rights are an attack on democracy. Example (d) objectifies women through a hypersexualized portrayal and prescribes strategies on how to attract or seduce them.}
    \centering
    \small
    \setlength{\tabcolsep}{4pt}  
    \renewcommand{\arraystretch}{1.2} 

    \begin{subfigure}{0.2\textwidth}
        \centering
        \includegraphics[width=0.6\linewidth, height=2cm]{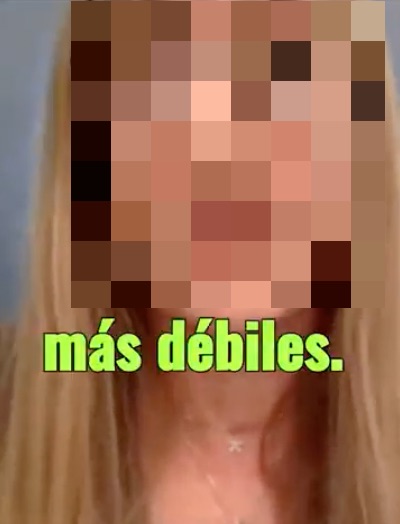}
        \caption{Stereotypes}
    \end{subfigure}
    \hfill
    \begin{subfigure}{0.2\textwidth}
        \centering
        \includegraphics[width=0.6\linewidth, height=2cm]{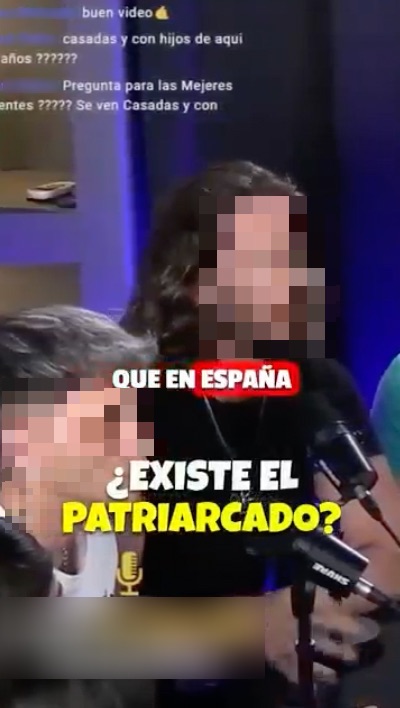}
        \caption{Inequality}
    \end{subfigure}
    \hfill
    \begin{subfigure}{0.2\textwidth}
        \centering
        \includegraphics[width=0.6\linewidth, height=2cm]{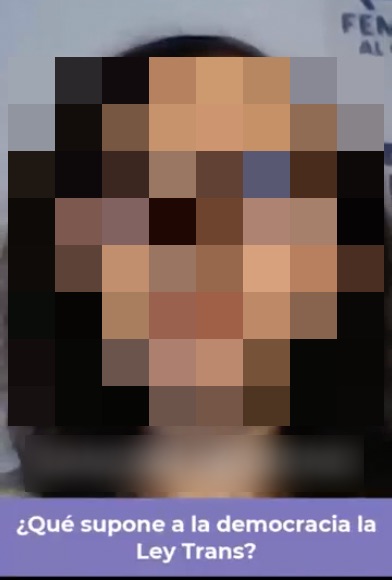}
        \caption{Discrimination}
    \end{subfigure}
    \hfill
    \begin{subfigure}{0.2\textwidth}
        \centering
        \includegraphics[width=0.6\linewidth, height=2cm]{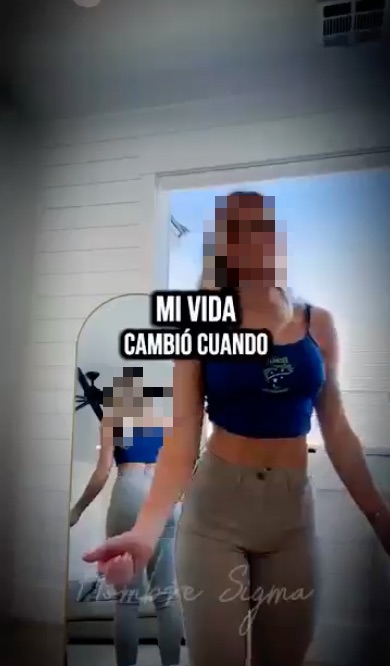}
        \caption{Objectification}
    \end{subfigure}

    \vspace{2mm} %

    \resizebox{\textwidth}{!}{
    \begin{tabular}{lp{7cm}p{7cm}} %
        \toprule
        \textbf{Sexist type}  & \textbf{Original text} & \textbf{English translation} \\
        \midrule
        Stereotypes (a) &  En todo lo que conocemos de la historia, las sociedades han organizado una forma que podríamos llamar patriarcal, en el sentido que han sido los hombres los que han tenido el poder exterior, no doméstico. 
        & Throughout all that we know of history, societies have been organized in a way that could be described as patriarchal, in the sense that it has been men who have held external, rather than domestic, power. 
        \\
        Inequality (b) & ¿Cómo puede ser que en España vivimos en un sistema patriarcal si hay más de 500 leyes, iniciativas y medidas que discriminan al hombre respecto a la mujer? & How can it be that in Spain we live in a patriarchal system if there are more than 500 laws, initiatives, and measures that discriminate against men compared to women? \\  
        Discrimination (c) & Esta es una ley borlaza, una ley restrictiva de todos los derechos y libertades. La mal llamada ley trans es un grave atentado contra el Estado de Derecho y la democracia. 
        & This is a sham law, a restrictive law that limits all rights and freedoms. The so-called “Trans Law” is a serious attack on the rule of law and democracy. 
        \\ 
        Objectification (d) & No hay que rogar ni mostrar debilidad. No hay que endiosar a las mujeres. Hay que enfocarse en metas.  & One should not beg or show weakness. One should not idolize women. One should focus on goals.
\\ 
        \bottomrule
    \end{tabular}
    }
    \label{tab:table_sexism_examples}
\end{table*}

\subsection{Sexist Content}
Sexist content refers to prejudice or discrimination based on sex, sexual orientation and gender identity. Sexism is classified into four categories, which are not mutually exclusive. Each instance of sexist content may fall under one or more of these categories. 

\subsubsection{Stereotypes}
Stereotypes are conceptualized in two ways: (i) as content that prescribes a set of behaviors or attributes that women and men are expected to exhibit in order to conform to gender roles; and (ii) as content that describes a set of supposed inherent traits that distinguish men from women \cite{samory2021call}.

\subsubsection{Denying Inequality and Rejection of Feminism}
Content that claims there are no longer inequalities between men and women and/or criticizes feminism \cite{samory2021call}. This category is conceptualized within the theoretical framework of Neo-sexism \cite{tougas1995neosexism}, which can manifest in three main forms: (a) denial of the existence of discrimination against women, (b) resentment towards complaints about discrimination, and (c) resentment towards special “favors” for women \cite{martinez2010predicting}.
This type of sexism also contributes to the spread of gendered disinformation \cite{gehrke2025gendered}, which relies on fake or manipulated studies to undermine women’s credibility \cite{tzvetkova2024applying}.

\subsubsection{Discrimination}
Content that discriminates against individuals based on their sexual orientation or gender identity, and that denigrates the LGBTQ+ community \cite{chakravarthi2024detection}. We observed that, in BitChute videos, creators often use conspiracy theories \cite{douglas2019understanding} to perpetuate discrimination against the LGBTQ+ community by referring to powerful elites who allegedly conspire to dismantle the traditional family structure and replace sexual identity with socially constructed gender roles. These findings are consistent with previous studies that identified BitChute as a powerful platform for disseminating conspiracy theories \cite{trujillo2020bitchute, mahl2022platformization, mahl2024conceptualizing}.

\subsubsection{Objectification}
Content that represents women as physical objects valued primarily for their utility to others \cite{szymanski2011sexual}. Objectification occurs when a woman’s body, or a part of it, is separated from her identity and is viewed as an object, shaped by the gaze and desire of the male. Table \ref{tab:table_sexism_examples} includes examples for each type of sexism.

\subsection{Non-sexist content}
Non-sexist content may fall under the following categories, which are not mutually exclusive. All of them are relevant for understanding how creators use social media platforms to challenge and report sexist behavior. 

\subsubsection{Counter-speech}
Counter-speech includes content that opposes, rejects, and critiques sexist behavior. For example,
\textit{Instead of responding with “not all men” to defend that they are the exception, men need to start listening to us.}

\subsubsection{Reported sexism}
Reported sexism includes the reporting of sexist situations in the first or third person. For example, \textit{He began to isolate me [...]. I used to have an athletics team that ended. My friendships also ended, because I had become a housewife.}


\subsubsection{Neither of the categories} Non-sexist content that does not fall into the previous two categories.

\subsection{Rhetorical devices}
Our taxonomy also incorporates annotations of the rhetorical devices of irony and humor. Accounting for figurative language enables a more detailed analysis of implicit expressions of sexism and facilitates the identification of irony and humor as a strategy for counter-speech.

\subsubsection{Irony}
In classical rhetoric, irony is a figure of speech where the words of the speaker convey the opposite of their intended meaning \cite{wilson1992verbal}. In this work, we also include sarcasm in the conceptualization of irony, following Gibbs’ classification, which defines sarcasm as a subtype of irony \cite{gibbs2000metarepresentations}. 

Sarcasm is characterized by the expression of an explicitly negative and critical attitude towards a person or an event. The following examples reflect the use of sexist (a) and non-sexist (b) irony:

\begin{enumerate}[label=\alph*)] 
    \item \textit {It is time to acknowledge that men hold social and legal privileges over women. 
    [...] For the same crime committed by a woman, men receive six additional years of punishment and are twice as likely to be incarcerated.} 
    \item \textit {Do not let your son play with dolls,
as there is a common belief that this might make him homosexual. Brooms, swaps, dolls – these are considered girls’ toys. This is what girls are taught from an early age.}
\end{enumerate}

\subsubsection{Humor}
Defined by the presence of amusing effects, such as laughter or a sense of well-being. In the context of sexist content, humor is often used to mock a target group, such as women or individuals who do not conform to the heterosexual and cisgender model. The following examples illustrate the use of sexist (a) and non-sexist (b) humor:

\begin{enumerate}[label=\alph*)] 
 \item \textit{You’re not crazy; you’re ecosexual.
And when you were walking through the park, you’d see that tree and think, “What a trunk it has.”
All you need to do is respect the plants, always get their consent,
and don’t forget to carry your safe-sex contract.}
 \item \textit{I think the most appropriate question would be to ask when they realized they are gay. Phrased literally, it might suggest that one could eventually become gay, or as if it were some kind of subscription that you pay for, with an exact start date. “Excuse me, I’d like to renew my homosexuality service.”
It doesn’t work that way.}
\end{enumerate}

\begin{figure}[h!]
    \centering
    \caption{Distribution of sexist and non-sexist categories across Part 1 and Part 2.}
    \begin{subfigure}[b]{0.45\textwidth}
        \centering
        \includegraphics[width=\textwidth]{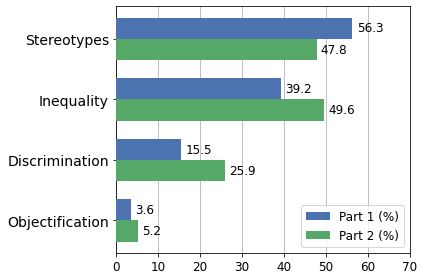}
        \caption{Sexist categories}
        \label{fig:sexist_types}
    \end{subfigure}
    \hfill
    \begin{subfigure}[b]{0.45\textwidth}
        \centering
        \includegraphics[width=\textwidth]{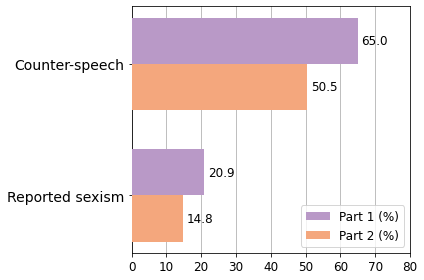}
        \caption{Non-sexist categories}
        \label{fig:non_sexist_types}
    \end{subfigure}
    \label{fig:comparison}
\end{figure}

\subsection{Annotation Process}
\label{sec:annotation}
The taxonomy was developed through a fine-grained annotation process, which is more challenging than binary annotation for several reasons. The use of multiple labels requires a deeper and more critical examination of the content, as annotators must consider several overlapping categories. Annotators may also experience information overload due to the large number of categories and subcategories involved \cite{guest2021expert}. Moreover, unconscious social biases among annotators can affect annotation quality \cite{zeinert2021annotating, grazia2025mused}, leading to interpretations of the guidelines that deviate from those intended by their authors. To improve the quality of the annotations with the fine-grained labels, we relied on expert annotators with prior experience in abusive language annotation and academic backgrounds related to sexism detection. They received specific training on the fine-grained classification task, including the use of the taxonomy (see Section~\ref{sec:taxonomy}) and the annotation guidelines (see Appendix~\ref{app:annotation}). We also engaged in iterative discussions about the use and appropriateness of fine-grained labels. The meetings were moderated by a facilitator with expertise in sexism on social media. Six annotators were involved in the task. The team included native speakers of both Peninsular and Latin American Spanish, ensuring greater cultural diversity than relying on a single dominant variety.\footnote{See Table~\ref{tab:annotators_profiles} in Appendix~\ref{app:annotation} for demographic details.} The annotation task was organized by splitting the initial group of annotators into two teams, each assigned to half of the dataset. Both teams were balanced in terms of gender and age, comprising one annotator who identifies as a man and two as women, with one member over 30 and the other two under 30. Following the annotation procedure introduced in \citet{grazia2025mused}, the process was carried out in three steps: first, annotating text transcripts; second, annotating audio; and finally, annotating videos (covering all modalities). Each team annotated the text transcripts and videos, while the other team annotated the audio. This multi-level annotation approach was applied to \textbf{Part 1} (P1), containing TikTok and BitChute videos sourced from MuSeD, using fine-grained labels,\footnote{MuSeD already provides binary labels. See Section~\ref{sec:data_sources}.}  
and to \textbf{Part 2} (P2), containing YouTube Shorts, using both binary and fine-grained labels. For Part 2, we further enriched the audio annotations by categorizing sexist types, thereby enabling a more comprehensive comparison across modalities. At each annotation step, annotators were able to select the specific temporal span of the fine-grained labels for text, audio, and video. This selection was important for identifying the segments in which multiple labels appear within the same item.

\begin{table*}[t]
\centering
\small
\caption{Distribution of Irony and Humor across FineMuSe.}
\begin{tabular}{lcccc}
\toprule
 & \multicolumn{2}{c}{\textbf{P1: TikTok \& BitChute}} & \multicolumn{2}{c}{\textbf{P2: YouTube Shorts}} \\
\cmidrule(lr){2-3} \cmidrule(lr){4-5}
\textbf{Type} & \textbf{Total} & \textbf{\%} & \textbf{Total} & \textbf{\%} \\
\midrule
Sexist            &  &  &  &  \\
\quad Irony & 5 & 2.4\% & 5  & 2.5\% \\
\quad Humor  &  7 & 3.4\% &  0 & 0\% \\
Non-sexist    &  &  &  &  \\
\quad Irony & 13 & 6.3\% &  10 & 4.3\% \\
\quad Humor &  10 & 4.9\% &  1 & 0.5\% \\
\bottomrule
\end{tabular}
\label{tab:irony_humor}
\end{table*}

\section{Dataset Distribution and Inter-Annotator Agreement} 
\label{sec:data_analysis}

\subsection{Categories Distribution}
The final dataset contains 828 videos distributed across three platforms: TikTok (367) and BitChute (33) for Part 1, and YouTube Shorts (428) for Part 2. FineMuSe shows a balanced distribution between sexist and non-sexist content, with 48.5\% of the videos labeled as sexist in Part 1 and 54.2\% in Part 2. Figure~\ref{fig:comparison} shows the distribution of sexist and non-sexist labels based on video-level annotations, which integrate all available modalities. Label counts were obtained using a majority-vote approach, whereby if at least two annotators selected a category, the final label was set to 1. In Figure~\ref{fig:sexist_types}, we observe that the categorization of sexist content is distributed differently across platforms, with the highest prevalence of Stereotypes in Part 1 and Inequality in Part 2. 
The limited presence of Objectification across FineMuSe suggests that this type of content is relatively scarce in the dataset. This may be due to stricter content moderation policies adopted by the platforms, which frequently ban material related to nudity and the sexualization of women’s bodies \cite{are2023autoethnography}.  
We also analyzed the distribution of non-sexist labels across platforms. Figure~\ref{fig:non_sexist_types} shows that both Counter-speech and Reported Sexism are more prevalent on TikTok than on YouTube Shorts. This indicates that content criticizing sexist beliefs against women, non-binary identities, and the LGBTQ+ community is more common on this platform. Moreover, content reporting experiences of sexism, whether directly experienced or witnessed, appears to have greater visibility on TikTok. We did not include BitChute as a platform for Counter-speech or Reported sexism, since 93.94\% of the videos in Part 1 were labeled as sexist \cite{grazia2025mused}. Finally, we analyzed the distribution of Irony and Humor, both sexist and non-sexist. Table~\ref{tab:irony_humor} shows that rhetorical devices are represented scarcely in the dataset, the most prevalent being non-sexist Irony, with only 13 instances (6.3\%) in Part 1. 

\subsection{Inter-Annotator Agreement}

\begin{table*}[t]
\centering
\small
\caption{Fleiss' Kappa values for the \textbf{Video} modality across dataset parts, annotation teams, and categories. Highest values per category are in \textbf{bold}. 
P1 = Part~1, P2 = Part~2.}
\begin{tabular}{lccc}
\toprule
\textbf{Category} & \textbf{Team} & \textbf{P1} & \textbf{P2} \\
\midrule
\multirow{2}{*}{Stereotypes}       & Team 1 & 0.52  &  0.50 \\
& Team 2 & \textbf{0.64}  &  0.50 \\
\multirow{2}{*}{Inequality}    & Team 1 &  0.57 &  0.72 \\
& Team 2 & 0.40  & 0.45  \\
\multirow{2}{*}{Discrimination}        & Team 1 & 0.54  &  \textbf{0.73}\\
& Team 2 & 0.60  &  0.49 \\
\multirow{2}{*}{Objectification}   & Team 1 & 0.28  &  0.49 \\
& Team 2 &  -0.01  &  0.30 \\
\midrule
\multirow{2}{*}{Counter-speech}    & Team 1 & \textbf{0.55}  &  0.38 \\
& Team 2 & 0.54  &  \textbf{0.39} \\
\multirow{2}{*}{Reported Sexism}   & Team 1 & 0.46 &  0.09 \\
& Team 2 & 0.28  &  0.34 \\
\midrule
\multirow{2}{*}{Sexist Irony}    & Team 1 & 0.37  & 0.06  \\ & Team 2 & 0.04  & 0.18  \\\multirow{2}{*}{Non-sexist Irony}   & Team 1 & 0.28  &  -0.00 \\ & Team 2 & 0.11 & -0.01  \\\multirow{2}{*}{Sexist Humor}    & Team 1 & 0.29  & \textbf{0.38}  \\& Team 2 & \textbf{0.42}  &  0.09 \\\multirow{2}{*}{Non-sexist Humor}   & Team 1 & 0.39  & 0.11  \\& Team 2 & 0.11  &  -0.00 \\
\bottomrule
\end{tabular}
\label{tab:kappa_video_categories}
\end{table*}

\begin{figure}[H]
    \centering
    \caption{Fleiss' Kappa values for sexist categories across Text and Video modalities in Part 1 and Part 2.}
    \begin{subfigure}{0.48\textwidth}
        \centering
        \includegraphics[width=\linewidth]{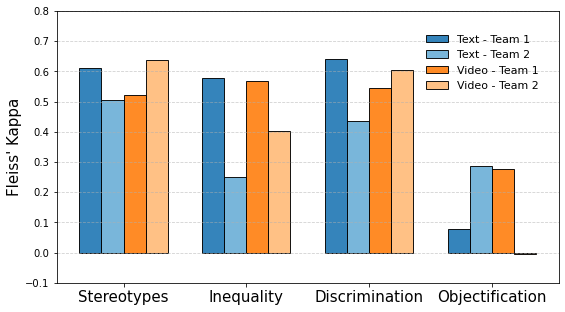}
        \caption{Part 1}
        \label{fig:1_kappa_sexism}
    \end{subfigure}
    \hfill
    \begin{subfigure}{0.48\textwidth}
        \centering
        \includegraphics[width=\linewidth]{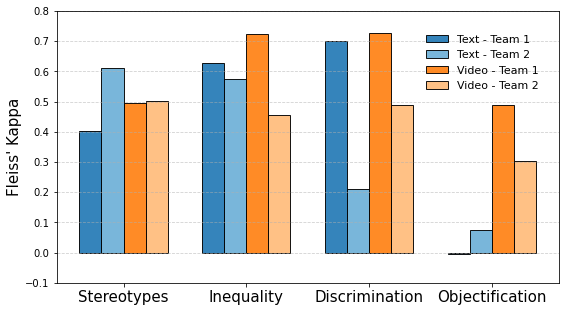}
        \caption{Part 2}
        \label{fig:2_kappa_sexism}
    \end{subfigure}
    \label{fig:kappa_sexism_combined}
\end{figure}

To assess the level of agreement among annotators, we used Fleiss’ Kappa \cite{fleiss1971measuring}. For the first level of binary classification, Fleiss’ Kappa for each team and modality is as follows: 0.69/0.67 for Text, 0.68/0.81 for Audio, and 0.81/0.67 for Video. The results were slightly lower than those obtained for Part 1, which were 0.72/0.72 for Text, 0.74/0.82 for Audio, and 0.83/0.85 for Video \cite{grazia2025mused}. The lower scores suggest that annotating \textit{shorts} is even more challenging than annotating TikTok and BitChute content. However, the obtained values remain higher than those reported for existing sexism datasets. \citet{arcos2024sexism} reported a Kappa of 0.50 for sexism detection on videos, indicating moderate agreement. In comparison, our Kappa values range from substantial (0.61–0.80) to almost perfect agreement (0.81–1) \cite{landis1977measurement, artstein2008inter}. Furthermore, we computed Fleiss' Kappa for the level-two categories, including fine-grained labels of sexist and non-sexist types for the video, given that it includes all the modalities. Table~\ref{tab:kappa_video_categories} shows the breakdown of Kappa scores per category for Part 1 and Part 2 of FineMuSe. The highest agreement was achieved on Stereotypes (k=0.61) in Part 1 and on Discrimination (k=0.73) in Part 2. Among non-sexist types, the highest agreement was observed for Counter-speech both in Part 1 (k=0.55) and Part 2 (k=0.39). Fleiss' Kappa was also computed for the Irony and Humor categories, including both their sexist and non-sexist subtypes. The low agreement observed for Irony and Humor, ranging from poor (< 0) to moderate (0.41–0.60), highlights the difficulty of reaching a consensual interpretation, as their perception is intrinsically subjective and varies between individuals depending on demographic factors (e.g. age and sex), as well as social and cultural influences \cite{frenda2023does}. Finally, we compared the Fleiss’ Kappa values for the text and video modalities, as these represent the two extremes of available information: minimal (text) and maximal (video). As shown in Figure~\ref{fig:1_kappa_sexism}, in Part 1, agreement was higher for the video modality in the Stereotypes, Inequality, and Discrimination categories, but only for Team 2. In contrast, in Part 2 (see Figure~\ref{fig:2_kappa_sexism}), agreement was higher for the video modality across all categories for Team 1. These results highlight the inherent difficulty of multilabel classification tasks, even when annotators have access to the full multimodal content.\footnote{See Appendix~\ref{app:iaa}
for a comparison of inter-annotator agreement (IAA) across modalities and labels. Tables~\ref{tab:fleiss_kappa_modalities_reduced} and~\ref{tab:fleiss_kappa_meta_discourse_nosexist} illustrate the IAA values for Part~1, while Tables~\ref{tab:fleiss_kappa_part2_full} and~\ref{tab:fleiss_kappa_part2_meta} show those for Part~2.}


\subsection{Analysis of Regional Spanish varieties}
\label{subsec:Spanish_analysis}
Considering the dialectical heterogeneity of our dataset, we analyzed the distribution of the regional varieties of Spanish. 
The assignment of each video to a Spanish variety was based on three key dimensions: prosodic features, such as intonation and stress patterns; lexical features, including the use of specialized terminology and vocabulary; and cultural markers, for instance, the incorporation of \textit{Spanglish}, which arises from the interaction between English and Spanish. 
To perform this analysis, we first considered two broad linguistic macro-areas: European Peninsular (Spain) and Latin American Spanish. Subsequently, we conducted a more detailed analysis of Latin American Spanish, dividing it into linguistic macro-areas corresponding to the following regional varieties \cite{martinez2025spanish}: Andean (Bolivia, Ecuador, and Peru), Antillean (Cuba, the Dominican Republic, and Puerto Rico), Chilean, Continental Caribbean (Colombia and Venezuela), Mexican and Central American (Costa Rica, El Salvador, Guatemala, Nicaragua, Panama, and Mexico), and Rioplatense (Argentina, Paraguay, and Uruguay). Table~\ref{tab:spanish_varities} illustrates the distribution of regional Spanish varieties, showing that Latin American Spanish constitutes the largest portion of the dataset. It was not always possible to identify the general linguistic macro-areas due to insufficient information, for example, in videos that used music, robotic or prerecorded voices. Among Latin American varieties, Mexican and Central American was the most frequently used across all the FineMuSe dataset. Mixed cases also show a prevalence of Latin American variety coexistence. We also examined the most frequent Latin American variety combinations, finding that, in Part 1, the most common pairs are Mexican and Central American with Continental Caribbean and Andean (seven cases each), while, in Part 2, the most frequent is Andean with Rioplatense (six cases).  


\begin{table*}[t]
\centering
\small
\caption{Distribution of regional Spanish varieties across dataset parts.}
\begin{tabular}{lcccc}
\toprule
 & \multicolumn{2}{c}{\textbf{P1: TikTok \& BitChute}} & \multicolumn{2}{c}{\textbf{P2: YouTube Shorts}} \\
\cmidrule(lr){2-3} \cmidrule(lr){4-5}
\textbf{Type} & \textbf{Total} & \textbf{\%} & \textbf{Total} & \textbf{\%} \\
\midrule
\textbf{Broad Macroareas} & & & & \\
\quad European Peninsular & 72 & 18\% & 110 & 25.7\% \\
\quad Latin American Spanish & 244 & 61\% & 257 & 60\% \\
\midrule
\textbf{Latin American Varieties} & & & & \\
\quad Andean & 30 & 12.3\% & 34 & 13.2\% \\
\quad Antillean & 6 & 2.5\% & 15 & 5.8\% \\
\quad Chilean & 6 & 2.5\% & 9 &  3.5\% \\
\quad Continental Caribbean & 27 & 11.1\% & 45 & 17.5\% \\
\quad Mexican and Central American  & 134 & 54.9\% & 117 & 45.6\% \\
\quad Rioplatense  & 41 & 16.8\% & 37 & 14.4\% \\
\midrule
\textbf{Mixed} & 42 & 10.5\% & 21 & 4.9\% \\
\quad European-Latin American & 11 & 2.75\% & 8 & 1.87\% \\
\quad Variety Coexistence & 27 & 6.75\% & 12 & 2.80\% \\
\quad Other & 4 & 1\% & 1 & 0.23\% \\
\textbf{Unclassified} & 42 & 10.5\% & 40 & 9.3\% \\
\bottomrule
\end{tabular}
\label{tab:spanish_varities}
\end{table*}


\section{Assessing Multimodal Models for Fine-Grained Sexism Detection} 
\label{sec:assessing_mm_models}

We frame fine-grained sexism detection as a hierarchical and \textit{multi-label classification problem}, following the taxonomy proposed in Section~\ref{sec:taxonomy}. In contrast to binary formulations that simply distinguish between \textit{sexist} and \textit{not sexist} content, our task requires first determining whether a social media post is sexist and, if so, identifying its specific \textit{type(s) of sexism}. In these experiments, we focus on the sexist categories, since the core objective of this work is to characterize and model nuanced expressions of sexist behavior. The label space for \textit{sexist} content consists of four non-exclusive classes, each capturing a distinct manifestation of sexism as described in Section~\ref{sec:taxonomy}. This formulation introduces additional complexity compared to binary classification, as sexist expressions frequently overlap, co-occur, or remain ambiguous, requiring a more context-sensitive and fine-grained interpretation.

\subsection{Models} 
In our experiments, we use a diverse set of models to capture different modalities and representation capabilities. 
We first evaluate \textbf{large language models (LLMs)} that process only textual input, including Llama-3-8B-Instruct, Llama-3-70B-Instruct \citep{dubey2024llama}, Qwen2.5-3B-Instruct \citep{yang2024qwen2}, Qwen2.5-32B-Instruct, Salamandra-7B-Instruct \citep{gonzalez2025salamandra}, Gemini-2.0-Flash \citep{team2024gemini}, GPT-4o \citep{hurst2024gpt}, and Claude~3.7 Sonnet \citep{anthropic2024claude}. 
These models serve as strong baselines and allow us to assess the extent to which sexism can be detected from transcripts and captions alone. To incorporate visual information, we further experiment with \textbf{multimodal LLMs} including Gemini-2.0-Flash (V+L), GPT-4o (V+L), and Claude~3.7 Sonnet (V+L), which integrate both text and images. 
This enables the detection of cues that may not be explicitly verbalized but are conveyed visually, such as gestures or visual stereotypes. Finally, we include Gemini-2.0-Flash (Video), which can process videos directly. 
Together, this combination of text-only and vision--language models provides a comprehensive evaluation setup that reflects the different ways sexism can be manifested and detected in social media video content.

\subsection{Experimental Setup}

We use a zero-shot prompting strategy, without fine-tuning or in-context examples, to simulate a realistic deployment scenario. We also instruct the models to provide a justification alongside each classification. 
All prompts enforce a strict JSON output format, ensuring consistency and enabling automated extraction and evaluation of the predicted sexism categories. 
Generations are produced with the temperature set to~0, yielding deterministic outputs and facilitating reproducible evaluation. For readability, we present the English translation of the video prompt in Figure \ref{fig:prompt-video-en} in Appendix~\ref{app:prompts}.
The original Spanish prompts, which were used in all experiments (for both text-only and video), are also provided in Appendix~\ref{app:prompts}.
\paragraph{Data Processing}
To ensure comparability with prior work, we adopt the pre-processing pipeline introduced in the MuSeD dataset \citep{grazia2025mused} without modification. This includes removing timestamps from transcripts to ensure models focus on the video content rather than metadata, 
extracting frames at evenly spaced intervals, capping video length at 80 seconds, allocating between two and 10 frames per video depending on duration, and filtering out black frames (commonly found in TikTok transitions) by detecting low pixel intensity. These steps ensure that only meaningful textual and visual information is retained for model input.

\paragraph{Evaluation Protocol} 
Results are presented for two parts of our dataset to enable comparison with previous work.
\textbf{Part~1} corresponds to the MuSeD dataset \citep{grazia2025mused}, which consists of videos from TikTok and BitChute and has been augmented with our fine-grained sexism labels.
\textbf{Part~2} contains additional data collected from YouTube Shorts (see 
Figure~\ref{fig:comparison} for an overview of the dataset distribution).
In both parts, we evaluate models on the binary sexism detection task, as well as on the fine-grained classification task. All evaluations are conducted against labels assigned by annotators who had access to the full video (both text and visual modalities; see Section~\ref{sec:annotation}). This ensures a consistent reference across text-only and multimodal models, reflecting the multimodal context in which content was originally interpreted. 

\paragraph{Evaluation Metrics} 
Following the MuSeD benchmark \citep{grazia2025mused}, we report \emph{Accuracy} for the binary sexism detection task to align with prior work. 
For the fine-grained multi-label task, our main metric is \emph{Macro F1}, as it balances performance across classes of different frequencies. 
We also report per-class F1 scores to give a detailed view of model strengths and weaknesses across the sexism categories. Since models occasionally fail to produce valid outputs (e.g., malformed JSON or refusals) and may differ in which instances they classify as sexist in the binary step, we include additional metrics to capture these behaviors: 
\textbf{Accuracy}, computed over all test instances with invalid or missing outputs counted as incorrect; 
\textbf{Valid Accuracy}, restricted to predictions that were validly formatted; 
\textbf{Macro F1}, computed over all ground-truth sexist instances, treating missed or invalid predictions as negatives in the multilabel setting; 
\textbf{Valid Macro F1}, restricted to cases where the model predicted the instance as sexist in the binary step and produced a valid fine-grained output; 
\textbf{Failure Rate (FR)}, the proportion of sexist instances missed because the model failed to detect them at the binary step and therefore provided no fine-grained labels; 
and \textbf{Format Error Rate (FoER)}, the proportion of outputs that were invalid and could not be parsed.



\begin{table}[t]
\centering
\small
\caption{Binary sexism detection results. 
Acc. = Accuracy over all samples. P1 = Part 1, P2 = Part 2. 
Format Error Rate (FoER) was negligible across models ($<0.07$ for all except Qwen2.5-32B-Instruct); 
full results with FoER and Valid Accuracy are reported in Appendix Table~\ref{tab:binary_appendix}. Best scores are in \textbf{bold}.}
\begin{tabular}{lcc}
\toprule
\textbf{Model} & \textbf{P1 Acc.} & \textbf{P2 Acc.} \\
\midrule
Random & 50.0 & 50.0 \\
\midrule
\multicolumn{3}{l}{\textit{Smaller LLMs}} \\
Llama-3-8B-Instruct & 61.8 & 67.0 \\
Qwen2.5-3B-Instruct & 61.0 & 66.6 \\
Salamandra-7B-Instruct & 55.0 & 60.1 \\
\midrule
\multicolumn{3}{l}{\textit{Larger LLMs}} \\
Llama-3-70B-Instruct & 83.3 & 75.1 \\
Qwen2.5-32B-Instruct & 72.3 & 73.8 \\
GPT-4o & 78.0 & 76.9 \\
Claude-3.7-Sonnet & 85.0 & 80.8 \\
Gemini-2.0-Flash & 72.5 & 72.4 \\
\midrule
\multicolumn{3}{l}{\textit{Multimodal LLMs}} \\
GPT-4o (V+L) & 79.5 & 78.1 \\
Claude-3.7-Sonnet (V+L) & \textbf{85.8}& \textbf{83.6} \\
Gemini-2.0-Flash (V+L) & 71.0 & 71.3 \\
Gemini-2.0-Flash (Video) & 68.3 & 72.7 \\
\bottomrule
\end{tabular}
\label{tab:results_binary}
\end{table}


\begin{table}[t]
\centering
\small
\caption{Fine-grained sexism detection results. 
F1 = Macro F1 (All), V-F1 = Valid Macro F1, FR = Failure Rate. Results are reported for Part 1 (P1) and Part 2 (P2). 
Format Error rates (FoER) are given in Appendix~\ref{app:results}. Best scores are in \textbf{bold}.}
\begin{tabular}{lcccccc}
\toprule
\textbf{Model} & 
\multicolumn{3}{c}{\textbf{P1}} & 
\multicolumn{3}{c}{\textbf{P2}} \\
\cmidrule(lr){2-4} \cmidrule(lr){5-7}
 & F1 & V-F1 & FR & F1 & V-F1 & FR \\
\midrule
Random     & 31.7 & 31.7 & 0.0 & 35.4 & 35.4 & 0.0 \\
\midrule
\multicolumn{7}{l}{\textit{Smaller LLMs}} \\
Llama-3-8B-Instruct       & 47.6 & 54.2 & 0.2 & 39.0 & 48.7 & 0.2 \\
Qwen2.5-3B-Instruct       & 25.1    & 26.6    & 0.1   & 27.2    & 29.3    & 0.2   \\
Salamandra-7B-Instruct    & 45.3    & 45.8    & 0.0   & 47.0    & 48.5    & 0.1   \\
\midrule
\multicolumn{7}{l}{\textit{Larger LLMs}} \\
Llama-3-70B-Instruct      & 39.8 & 42.2 & 0.0 & 41.1 & 50.6 & 0.2 \\
Qwen2.5-32B-Instruct      & 49.7 & 58.1 & 0.1 & 36.6 & 56.8 & 0.1 \\
GPT-4o                    & 62.1 & 64.9 & 0.1 & 63.2 & 70.5 & 0.2 \\
Claude-3.7-Sonnet         & 51.6 & 55.4 & 0.1 & 52.1 & 59.9 & 0.2 \\
Gemini-2.0-Flash          & 50.9 & 54.7 & 0.1 & 48.3 & 59.8 & 0.3 \\
\midrule
\multicolumn{7}{l}{\textit{Multimodal LLMs}} \\
GPT-4o (V+L)              & 65.3 & 68.1 & 0.1 & 67.4 & 73.5 & 0.1 \\
Claude-3.7-Sonnet (V+L)   & \textbf{67.5} & \textbf{68.2} & 0.0 & \textbf{72.8} & 76.3 & 0.1 \\
Gemini-2.0-Flash (V+L)    & 58.7 & 65.4 & 0.2 & 64.6 & \textbf{79.3} & 0.3 \\
Gemini-2.0-Flash (Video)  & 47.2 & 53.7 & 0.2 & 54.8 & 62.3 & 0.2 \\
\bottomrule
\end{tabular}
\label{tab:results-fg}
\end{table}


\subsection{Experimental Results} 

We present experimental results for the two parts of the dataset: Part~1, consisting of TikTok and BitChute videos from the MuSeD benchmark enriched with our fine-grained labels, and Part~2, consisting of additional data collected from YouTube Shorts (see Section~\ref{sec:data_collection}). Binary sexism detection results (Accuracy) are reported in Table~\ref{tab:results_binary}, fine-grained multi-label results (Macro F1) in Table~\ref{tab:results-fg}, and per-class F1 scores for each sexism category in Figure~\ref{fig:per_class_f1} and in Table~\ref{tab:fg_perclass} in Appendix~\ref{app:results}.

\paragraph{Impact of Model Scale} 
Model scale plays an important role in performance. Larger LLMs consistently outperform their smaller counterparts, both in binary accuracy and in fine-grained classification. For example, Llama-3-70B and Qwen2.5-32B clearly improve over their smaller versions, while GPT-4o and Claude-3.7-Sonnet achieve the strongest results overall. These findings confirm that larger models are better equipped to capture the subtle and diverse expressions of sexism, whereas smaller models often fail to generalize across categories and miss nuanced cases.

\paragraph{Binary Sexism Detection} 
Binary classification results in Table~\ref{tab:results_binary} show that both large LLMs and multimodal LLMs substantially outperform random baselines. Among text-only models, Claude-3.7-Sonnet achieves the highest accuracy (85.0 in P1, 80.8 in P2), followed closely by Llama-3-70B. Multimodal models yield further improvements, with Claude-3.7-Sonnet (V+L) achieving the strongest overall performance (85.8 in P1, 83.6 in P2). These results suggest that access to visual context provides complementary information, particularly in Part~2 (YouTube Shorts), where multimodal cues appear to boost performance. This complements the findings in \citet{grazia2025mused} for binary classification. Format Error Rate (see Table~\ref{tab:binary_appendix} in the Appendix) was negligible for almost all models ($<0.07$), ensuring that observed improvements reflect genuine model ability rather than formatting issues.

\paragraph{Fine-grained Sexism Detection} 
The fine-grained task proves substantially more challenging than binary detection as shown in Table~\ref{tab:results-fg}, with random baselines already reaching F1 scores of around 31--35 due to class imbalance. Smaller LLMs perform poorly, barely surpassing 40 in macro-F1, and frequently exhibit high failure rates. In contrast, larger models achieve more robust results: GPT-4o reaches 62.1 F1 in P1, while Claude-3.7-Sonnet (V+L) delivers the best performance in both P1 and P2 (67.6 and 72.9 F1). Multimodal models consistently outperform text-only counterparts, especially in Part~2, where visual context appears critical for handling the more diverse YouTube Shorts content. Nevertheless, even the strongest models show notable failure rates (up to 0.34 for Gemini-2.0-Flash V+L), underscoring the difficulty of reliably predicting fine-grained sexism categories.

\begin{figure}[t]
    \centering
    \includegraphics[width=\linewidth]{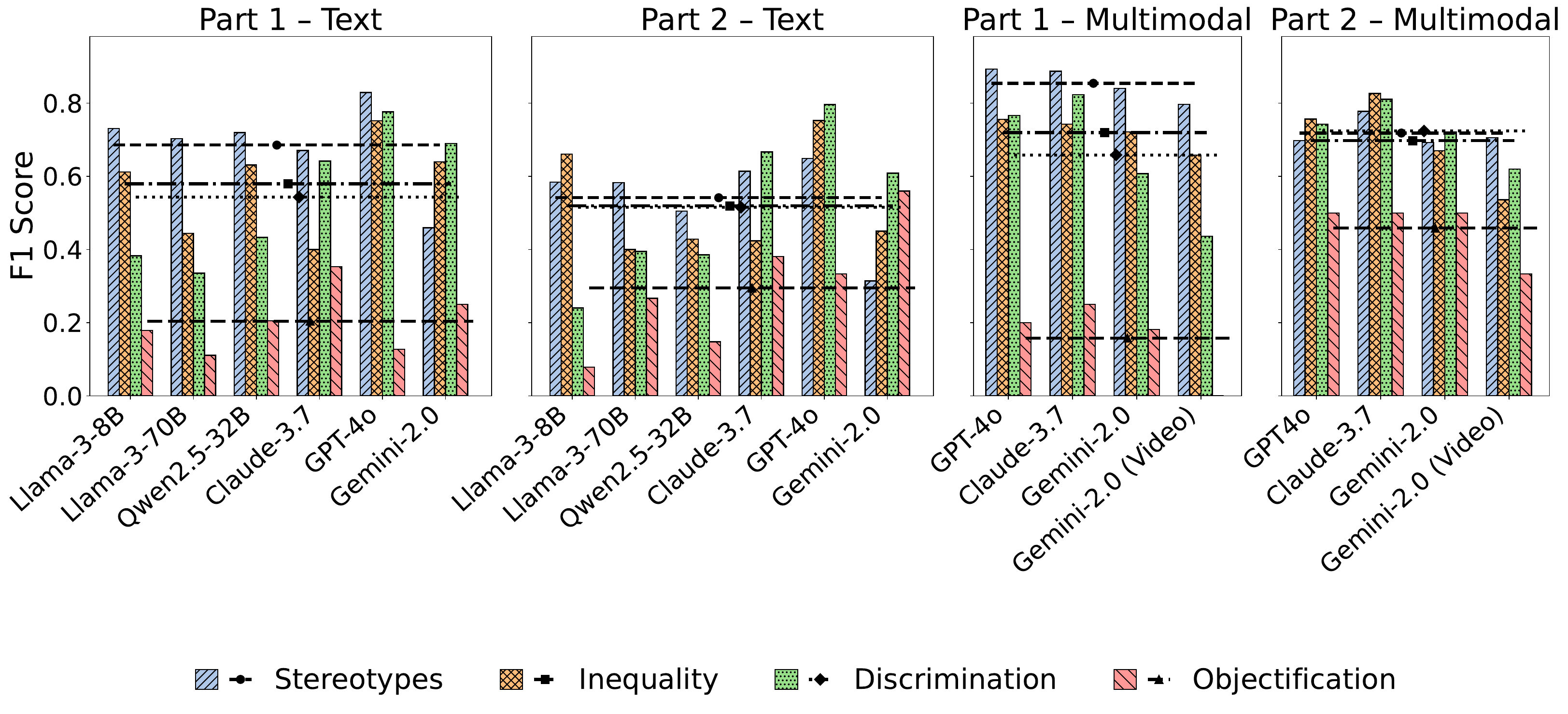}
   \caption{Per-class F1 scores for fine-grained sexism detection across text-only and multimodal models. 
    Results are split by dataset part (Part~1, Part~2). Horizontal dashed lines denote category averages across models.
    Multimodal models consistently outperform text-only models, especially in Part~2, though Objectification remains the hardest category across platforms.}
    \label{fig:per_class_f1}
\end{figure}

\paragraph{Per-class Performance} 
 Figure~\ref{fig:per_class_f1} shows that performance varies considerably across sexism categories. Stereotypes and Inequality are the easiest to detect, with multimodal models such as GPT-4o (V+L) and Claude-3.7-Sonnet (V+L) achieving F1 scores above 69 in both parts. Discrimination is more challenging but still shows strong results for multimodal models, e.g., Claude-3.7-Sonnet (V+L) at 82.4 F1 in P1 and 81.1 F1 in P2. Objectification is consistently the hardest category: in Part~1 most models remain below 30 F1, and only a few multimodal systems reach around 50 F1 in Part~2. This result likely reflects the scarcity of Objectification instances in the dataset. As shown in Section~\ref{sec:data_analysis}, this category appears far less often, partly due to stricter moderation of content involving nudity or sexualization \cite{are2023autoethnography}, which limits the prior knowledge models can draw on in a zero-shot setting and contributes to their lower performance.

\paragraph{Cross-Platform Differences} 
Model performance also differs across platforms. In Part~2 (YouTube Shorts), Inequality and Discrimination scores generally improve, often exceeding 70 F1 for the strongest multimodal models. At the same time, performance on Stereotypes declines by 10–15 points compared to Part~1. This indicates platform-specific differences in how sexism manifests: TikTok and BitChute content more often emphasizes stereotypical portrayals of gender roles, whereas YouTube Shorts contain more argumentative discourse around gender inequality and discrimination, where multimodal cues provide clearer benefits. Objectification remains the most difficult category across both parts, though multimodal models achieve modest improvements in Part~2 compared to their unimodal counterparts.

\begin{figure}[t]
    \centering
    \caption{Correlation matrices for human annotations and Claude-3.7 Sonnet (V+L) predictions, based on the co-occurrence of sexist types.}
    \label{fig:cor_matrix}

    \begin{subfigure}[t]{0.45\textwidth}
        \centering
        \includegraphics[width=\linewidth]{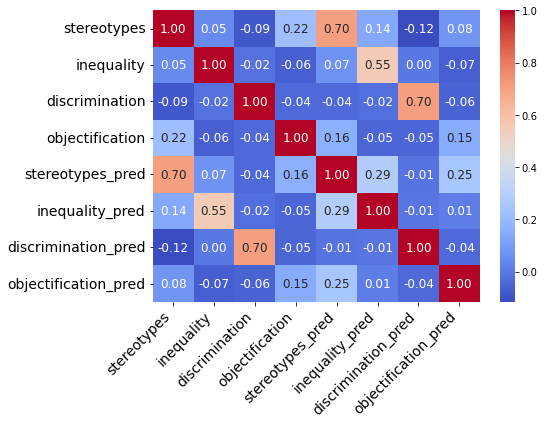}
        \caption{Part 1}
        \label{fig:cor_matrix_part1}
    \end{subfigure}\hfill
    \begin{subfigure}[t]{0.45\textwidth}
        \centering
        \includegraphics[width=\linewidth]{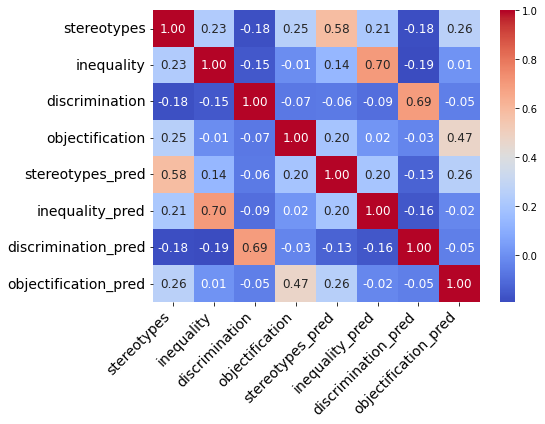}
        \caption{Part 2}
        \label{fig:cor_matrix_part2}
    \end{subfigure}
\end{figure}

\section{Human–Model Comparative Analysis}
In this section, we examine how closely models align with human annotators in both their predictions and their explanations. 

\subsection{Correlation Analysis}
\label{subsec:occurrence_errors}

We use the fine-grained labels to conduct an error analysis by examining the correlations between sexist categories assigned by human annotators and those predicted by Claude-3.7-Sonnet (V+L), the best-performing model for fine-grained detection. Figure~\ref{fig:cor_matrix} shows the Pearson correlation matrices for Part 1 (TikTok, BitChute) and Part 2 (YouTube Shorts). The upper-left 4×4 block of each matrix reports correlations among human annotations obtained via majority vote (see Section~\ref{sec:data_analysis}). In Part 1 (see Figure~\ref{fig:cor_matrix_part1}), the strongest positive correlation among human labels is observed between Stereotypes and Objectification (r = 0.22, p < 0.001), suggesting that gender stereotypes often co-occur with representations that portray women as objects rather than agents \cite{rollero2016effects}. The second block captures correlations between human annotations and model predictions. In Part 1, strong correlations are observed for Stereotypes (r = 0.70) and Discrimination (r = 0.70), indicating substantial alignment between model predictions and human judgments for these categories. In contrast, the correlation for Objectification is comparatively low (r = 0.15), suggesting that this category poses greater challenges for the model. For instance, in the example \textit{How to be funny and attract someone without looking ridiculous? My brother, a key skill to be funny and attractive is intentional misinterpretation}, the model correctly identifies the presence of Stereotypes in the verbal content but fails to detect Objectification conveyed through the visual modality. The bottom-right block reports correlations among the predicted labels of the model, reflecting how the model internally associates different categories independently of human agreement. In Part 1, the model exhibits a stronger positive association between Stereotypes and Inequality (r = 0.29) than that observed in human annotations, indicating that these categories co-occur more frequently in the model’s predictions than in human judgments. For Part 2 (see Figure~\ref{fig:cor_matrix_part2}), human annotations show that Stereotypes tend to co-occur with both Inequality (r = 0.23, p < 0.001) and Objectification (r = 0.25, p < 0.001). Correlations between human annotations and model predictions are high for Inequality (r = 0.70) and Discrimination (r = 0.69), while Objectification shows a moderate correlation (r = 0.47). An analysis of misclassified cases for Objectification reveals a pattern similar to that observed in Part 1: the model successfully detects Stereotypes expressed in verbal content but struggles to recognize Objectification conveyed visually. For example, in \textit{If you want to succeed in life, call everyone by their name, and speak only when necessary. Master a cold and calculating gaze, and do not let others intimidate you}, the verbal content prescribes normative masculine behavior (Stereotypes), while the visual content involves hypersexualization of the female body (Objectification), which the model struggles to predict. Finally, correlations among the model’s predicted labels in Part 2 reveal a positive relationship between Stereotypes and Inequality, consistent with the co-occurrence patterns observed in human annotations.

\subsection{Explanation Quality}
\label{subsec:just_quality}
To assess explanation quality, we compared justifications produced by Claude-3.7-Sonnet (V+L), the best performing model, with those written by human annotators. We selected 100 videos from FineMuSe, including all cases labeled with Irony and Humor (51; see Table \ref{tab:irony_humor}), and a proportional random sample, from Parts 1 and 2.
Three expert annotators each wrote justifications (35-50 words) for 35 videos and then rated both
human and model explanations, blinded to their source and ensuring annotators did not rate videos they had written themselves.\footnote{Table \ref{tab:comparison_explanations} presents a sample comparison of the model and the evaluators based on the three metrics.} 
Explanations were rated using a 5-point Likert scale across three metrics: (i) Relevance, does the explanation directly address the video and its classification label?; (ii) Informativeness: how much useful information does the explanation provide, rather than simply repeating content?; (iii) Groundedness, are the claims supported by the video? Explanations must only include claims that can be verified in the video, without introducing manipulative or fabricated content \cite{dziri2021evaluating}. Illustrative examples of explanations for sexist content are reported below. The first explanation is produced by the model, whereas the second is written by a human annotator: (i) \textit{The content ironically presents supposed `male privileges' which are actually disadvantages, minimizing structural inequalities faced by women and reinforcing gender stereotypes about work roles and family responsibilities}; (ii) \textit{The video presents sexist content as it questions male privilege, arguing that men occupy higher-risk positions, retire later, and often lose child custody cases}. We employed the Wilcoxon signed-rank test \cite{wilcoxon1945individual} to assess whether annotators systematically rated human explanations higher or lower than model-generated ones. 
For Relevance, only one annotator showed a statistically significant preference for human explanations; the other two gave slightly higher average scores to the model, but these differences were not statistically significant. For Informativeness, humans received higher average ratings in most cases, though none reached statistical significance. For Groudedness, no significant differences were observed.\footnote{Table \ref{tab:wilcoxon_results} reports p-values, averages, and medians.} Overall, these findings indicate that LLMs can produce explanations of comparable quality to human annotators. This aligns with prior work showing that models such as GPT-3 can generate fluent, informative, persuasive, and logically coherent explanations when appropriately prompted \cite{wang2023evaluating}. 
 
\section{Conclusion and Future Work}\label{sec:conclusion}
We introduced FineMuSe, a new multimodal dataset for sexism detection that provides fine-grained labels across text, audio and video. We conducted an extensive empirical study assessing a wide range of LLMs on both binary and fine-grained sexism detection. Our findings show that state-of-the-art multimodal LLMs are competitive with humans in identifying whether content is sexist. 
However, they struggle to capture co-occurring sexist types when these are conveyed through visual cues. Studying how humans and models use and prefer different forms of justification can support the development of more transparent and trustworthy moderation tools. Future work may further analyze how each modality contributes to detecting nuanced forms of sexism. Additionally, improving data collection for irony and humor would enable a deeper understanding of how rhetorical devices influence misclassification.

\clearpage



\appendix
\setcounter{section}{0}

\AppendixSection{Hashtags}\label{app:hashtags}

\#alfa, \#abortodebate, \#abortolegal, \#acosocallejero, \#alfaseductor, \#alfasenproceso, \#alpha, \#amorromantico, \#antifeminismo, \#antifeminista, \#atraccion, \#binarie, \#bisexual, \#brechadegenero, \#brechadegénero, \#brechasalarial, \#buscarpareja, \#consejoparamujeres, \#consejosparahombres, \#contralaviolenciadegenero, \#ContraLaViolenciaDeGenero, \#cosificacion, \#culturapatriarcal, \#debateprovida, \#derecho, \#derechosdelamujer, \#derechossexualesyreproductivos, \#desigualdaddegenero, \#desigualdaddegénero, \#discriminación, \#discriminacióndegénero, \#diversidadsexual, \#educacionsexual, \#empoderamiento, \#empoderamientofemenino, \#empoderamientomaterno, \#equidad, \#equidaddegenero, \#estereotipos, \#estereotiposdegenero, \#expresióndegénero, \#familia, \#feminicidio, \#feminidad, \#feminismo, \#feminismoactual, \#feminismocritico, \#feminismomoderno, \#feminismoradical, \#feminismos, \#feminismotoxico, \#feminismoymaternidad, \#feminista, \#feministas, \#feministaslatinoamericanas, \#feministasmodernas, \#feministastiktok, \#feministastoxicas, \#gay, \#genero, \#género, \#hembrismo, \#hembrismonoesfeminismo, \#heteronormatividad, \#heterosexual, \#hombre, \#hombrealfa, \#hombredealtovalor, \#hombres, \#hombresalfa, \#hombresalpha, \#hombresaltovalor, \#hombresdealtovalor, \#hombrestradicionales, \#hombresvaliosos, \#hombresvsmujeres, \#hombretoxicos, \#homofobia, \#homosexual, \#homosexualidad, \#identidad, \#identidaddegenero, \#identidadsexual, \#identidadsexualdeloshijos, \#ideologíadegénero, \#igualdad, \#igualdaddegenero, \#igualdadentrehombresymujeres, \#inclusion, \#juntasporlaigualdad, \#lasmujeressomosprioridad, \#lenguajeinclusivo, \#leytrans, \#lgbt, \#lgbtq, \#lgbtqia, \#lgtbiq, \#machismo, \#machismoonline, \#machista, \#machoalfa, \#machosalpha, \#marchasfeministas, \#masculinidad, \#masculinidades, \#masculinidadfragil, \#masculinidadfuerte, \#masculinidadtóxica, \#maternidad, \#maternidadreal, \#maternidadsinfiltro, \#matriarcado, \#micuerpoesmioyyodecido, \#misandria, \#misandrianormalizada, \#misandrica, \#misoginia, \#misoginiainteriorizada, \#monogamia, \#mujer, \#mujerdealtovalor, \#mujerdelirante, \#mujerempoderada, \#mujeres, \#mujeresaltovalor, \#mujeresdealtovalor, \#mujeresempoderadas, \#mujereslatinas, \#mujeresqueinspiran, \#mujeressolteras, \#mujerestoxicas, \#mujersuperior, \#mujertoxica, \#mujertrans, \#niunamenos, \#niñas, \#noalaviolenciadegenero, \#nobinario, \#noesno, \#noestassola, \#nomasviolencia, \#nosoyfeminista, \#noviostoxicos, \#nuevasmasculinidades, \#orientacionsexual, \#parejasenamoradas, \#parejastoxicas, \#paridaddegenero, \#paternidad, \#patriarcado, \#patriarcadoopresor, \#privilegio, \#privilegiomasculino, \#provida, \#providas, \#psicologiadepareja, \#psicologiafemenina, \#relacionesdepareja, \#relacionessanas, \#relacionestoxicas, \#relaciónsana, \#serfeminista, \#sermujer, \#serprovida, \#sexismo, \#sexo, \#sexualidad, \#soyfeminista, \#soymujerantifeminista, \#techodecristal, \#tiposdemujeres, \#toxicas, \#toxicasnovias, \#toxicidad, \#toxicos, \#trabajoremunerado, \#trans, \#transexual, \#transexualidad, \#transgénero, \#violencia, \#violenciadegenero, \#violenciadegeneronoexiste, \#violenciadegenerostop, \#violenciamachista, \#virilidad, \#yotecreo. 

\clearpage

\AppendixSection{Human Annotation}\label{app:annotation}

\begin{table}[H]
    \caption{Annotator profiles, including gender, age groups, languages, and areas of study/occupation.}
    \centering
    \small
    \begin{tabular}{ll}
        \toprule
        \multicolumn{2}{c}{\textbf{Annotator profiles}} \\  
        \midrule
        Gender          & 4 female, 2 male (6 total)  \\  
        Age            & 3 under 30, 3 over 30 \\  
        Native Language & 
        Mexican Spanish (1), 
        Peninsular Spanish (5)\\  Study/Occupation & Linguistics (4), Adjunct Professor (1), \\ & PhD student in Science Communication (1)
        \\
        \bottomrule
    \end{tabular}
    \label{tab:annotators_profiles}
\end{table}


\textit{Annotation Guidelines}. The focus of the annotation is to classify the content using binary labels (\textbf{sexist} or \textbf{non-sexist}), and then apply a \textbf{subcategorization} according to the selected label. Each subcategorization must be selected by highlighting the specific segment in which it appears. Content is considered sexist if it includes discrimination based on sex, sexual orientation, or gender identity. Whether the content is classified as sexist, one or more of the following categories must be selected. Different forms of sexism can coexist, so you may select multiple labels.

\vspace{0.5cm}
\textbf{1. Stereotypes}
This category includes content that formulates a descriptive set of properties that differentiate men and women (a) and/or formulates a prescriptive set of qualities that men and women are expected to exhibit in order to conform to gender roles (b). 

\begin{enumerate}[label=\alph*)] 
    \item Mi gente, mi gente, miren, disfruten de estos juguetes… porque las feministas pronto van a protestar que las niñas no deben jugar con esto. \textit{English translation}: My people, my people, look, enjoy these toys… because feminists will soon protest that girls shouldn’t play with them.
    \item ¿Qué hace un alfa cuando una mujer lo rechaza? Muchos hombres se lo toman muy personal y tratan de obtener una explicación. Nunca le pidas una explicación, porque te hará ver como un hombre necesitado. \textit{English translation}: What does an alpha do when a woman rejects him? Many men take it very personally and try to get an explanation. Never ask for an explanation, because it will make you look like a needy man.
\end{enumerate}

\vspace{0.5cm}
\textbf{2. Denying Inequality and Rejection of Feminism}
Content stating that there are no inequalities between men and women (a) and/or that opposes feminism (b).

\begin{enumerate}[label=\alph*)] 
    \item Es hora de admitirlo, los hombres tenemos privilegios sobre las mujeres, sociales y legales. Por eso me he tomado la molestia de buscar información verificada…\textit{English translation}: It’s time to admit it: men have privileges over women, both socially and legally. That’s why I took the time to gather verified information…
    \item Tienes a las feministas protestando por el heteropatriarcado, el privilegio masculino. Todo va a favor de los hombres. Nosotras vivimos oprimidas. \textit{English translation}: You have feminists protesting against the heteropatriarchy, male privilege. Everything works in favor of men. We live oppressed.
\end{enumerate}

\vspace{0.5cm}

\textbf{3. Discrimination} 

Content that discriminates based on sexual orientation (a), gender identity (b), and against the LGBTQ+ community (c).

\begin{enumerate}[label=\alph*)] 
    \item ¿Por qué hay más gays, lesbianas, bisexuales, transexuales en el mundo? Porque les vendieron la mentira de la identidad sexual para pertenecer a un grupo… \textit{English translation}: Why are there more gay, lesbian, bisexual, and transgender people in the world? Because they were sold the lie of sexual identity in order to belong to a group…
    \item La doctora Mcnamara afirma que sus puntos de vista se basan en la ciencia. Afirmar que el sexo se asigna al nacer no tiene base científica. Su lenguaje engaña especialmente a los niños. \textit{English translation}: Dr. McNamara claims that her views are based on science. Stating that sex is assigned at birth has no scientific basis. Her language is particularly misleading for children.
    \item Vamos a estar atentos a la instalación de la comisión del colectivo LGTBIQ+, que pretende instaurar normas que van en contra de la familia. \textit{English translation}: We will be closely monitoring the establishment of the LGTBIQ+ collective commission, which intends to introduce rules that go against the family.
\end{enumerate}

\vspace{0.5cm}
\textbf{4. Objectification} 

Content that represents women as objects to conquer (a) and/or judges them for their physical appearance (b).

\begin{enumerate}[label=\alph*)] 
    \item Tipos de mujeres que debes evitar. La que acaba de salir de una relación. Muchas veces no están listas emocionalmente. 
Algunas te pueden usar para olvidar o darle celos a su ex incluso, para llenar un vacío. \textit{English translation}: Types of women you should avoid. The one who has just gotten out of a relationship, often, they are not emotionally ready. Some may use you to forget their ex or even to make them jealous, to fill a void.
    \item Lo único que un hombre pide para amar a una mujer es que sea femenina, atractiva y fértil. \textit{English translation}: The only thing a man asks to love a woman is that she is feminine, attractive, and fertile. 
\end{enumerate}

\vspace{0.5cm}
\textbf{Non-sexist content}

Whether the content is labeled as \textbf{non-sexist}, it must be categorized as Counter-speech (a), Reported sexism (b) or None of the categories:

\begin{enumerate}[label=\alph*)] 
    \item  No hay una forma única de vivir la sexualidad. Cada persona debe ser aceptada por quien es. Las violencias y la discriminación atentan con la posibilidad de llevar una vida digna. \textit{English translation}: There is no single way to experience sexuality. Each person should be accepted for who they are. Violence and discrimination undermine the possibility of leading a dignified life.
    \item No había teléfono porque yo no puedo estar viendo con quién te vas a estar comunicando. No había un comedor porque yo solo quería comedor para que la gente viniera de metiche a opinar. \textit{English translation}:There was no phone because I can’t be keeping track of who you’re talking to. There was no dining room because I only wanted it so people would come in and meddle with their opinions.
\end{enumerate}

\newpage

\textbf{Irony and Humor}

A content could be reinforced using irony or humor. By irony, we mean a message that expresses a meaning opposite to what is said. It could convey a \textbf{sexist} (a) or \textbf{non-sexist} (b) meaning.

\begin{enumerate}[label=\alph*)] 
    \item El privilegio que tienen los hombres en la Argentina es que se jubilan 5 años después y mueren en promedio 7 años antes. Es decir, que ellos tienen oportunidad de seguir sintiéndose útiles para la sociedad, mientras que a nosotros nos tiran como trastos viejos y empezamos a cobrar la jubilación antes. \textit{English translation}: The privilege that men have in Argentina is that they retire five years later and, on average, die seven years earlier. In other words, they have the chance to keep feeling useful to society, while we are tossed aside like old junk and start collecting our pension earlier.
    \item No permitas que tu hijo juegue con muñecas. Hay altas posibilidades de que sea homosexual. Las escobas, los swaps, las muñecas son cosas de mujeres. Eso nos hicieron creer desde niñas. \textit{English translation}: Don’t allow your son to play with dolls. There’s a high chance he’ll be homosexual. Brooms, mops, dolls—those are things for women. That’s what they made us believe since we were little girls.
\end{enumerate}

By Humor, we refer to the presence of amusing effects, such as laughter or a sense of well-being. In sexist cases (a), it may be used to poke fun at a target group, such as women or people who do not conform to heterosexual and cisgender norms. In non-sexist contexts (b), it can be used to mock sexist beliefs and behaviors.

\begin{enumerate}[label=\alph*)] 
    \item Speaker A: Creo que soy lesbiana. 
    
    Speaker B: Pedro, no puedes ser lesbiana porque eres un hombre. 
    
    Speaker A: Pero si lo piensas bien, todo encaja. Yo no era una mujer atrapada en el cuerpo de un hombre. Soy una lesbiana atrapada en el cuerpo de un gay. 
    
    \textit{English translation}: Speaker A: I think I’m a lesbian. 
    
    Speaker B: Pedro, you can’t be a lesbian because you’re a man. 
    
Speaker A: But if you think about it, it all makes sense. I wasn’t a woman trapped in a man’s body; I’m a lesbian trapped in the body of a gay man.
    \item Imaginate siendo mujer. Caminar solo de noche en Argentina y sobrevibir. Tendría que ser reconocido como un logro para el CV. \textit{English translation}: Imagine being a woman. Walking alone at night in Argentina and surviving. It should be recognized as an achievement on your résumé.  
\end{enumerate}

\AppendixSection{Inter-Annotator Agreement}\label{app:iaa}

\begin{table}[H]
\centering
\caption{Fleiss' Kappa values across modalities (Text, Audio, and Video) for Part 1 on sexist categories.}
\label{tab:fleiss_kappa_modalities_reduced}
\begin{adjustbox}{max width=\textwidth}
\begin{tabular}{lcccccccc}
\toprule
\textbf{Modality} & \textbf{Team} & \textbf{Sexist} & \textbf{Stereotype} & \textbf{Inequality} & \textbf{Discrimination} & \textbf{Objectification} & \textbf{Irony} & \textbf{Humor} \\
\midrule
\multirow{2}{*}{\textbf{Text}} 
 & Team 1 & 0.72 & 0.61 & 0.58 & 0.64 & 0.08 & 0.20 & 0.41 \\
 & Team 2 & 0.72 & 0.51 & 0.25 & 0.44 & 0.29 & 0.29 & 0.32 \\
\midrule
\multirow{2}{*}{\textbf{Audio}} 
 & Team 1 & 0.74 & \ding{55} & \ding{55} & \ding{55} & 0.47 & 0.17 & 0.34 \\
 & Team 2 & 0.82 & \ding{55} & \ding{55} & \ding{55} & 0.50 & 0.48 & 0.39 \\
\midrule
\multirow{2}{*}{\textbf{Video}} 
 & Team 1 & 0.83 & 0.52 & 0.57 & 0.54 & 0.28 & 0.37 & 0.28 \\
 & Team 2 & 0.85 & 0.64 & 0.40 & 0.60 & -0.01 & 0.04 & 0.11 \\
\bottomrule
\end{tabular}
\end{adjustbox}
\end{table}

\begin{table}[H]
\centering
\caption{Fleiss' Kappa values across modalities (Text, Audio, and Video) for Part 1 on non-sexist categories.}
\label{tab:fleiss_kappa_meta_discourse_nosexist}
\begin{adjustbox}{max width=\textwidth}
\begin{tabular}{lcccccc}
\toprule
\textbf{Modality} & \textbf{Team} & \textbf{Counter-speech} & \textbf{Reported sexism} & \textbf{Irony} & \textbf{Humor} \\
\midrule
\multirow{2}{*}{\textbf{Text}} 
 & Team 1 & 0.53 & 0.39 & 0.41 & 0.37 \\
 & Team 2 & 0.10 & 0.27 & 0.32 & -0.01 \\
\midrule
\multirow{2}{*}{\textbf{Audio}} 
 & Team 1 & 0.47 & 0.34 & 0.17 & 0.41 \\
 & Team 2 & 0.50 & 0.44 & 0.48 & 0.22 \\
\midrule
\multirow{2}{*}{\textbf{Video}} 
 & Team 1 & 0.55 & 0.46 & 0.28 & 0.39 \\
 & Team 2 & 0.54 & 0.28 & 0.11 & 0.11 \\
\bottomrule
\end{tabular}
\end{adjustbox}
\end{table}

\begin{table}[H]
\centering
\caption{Fleiss' Kappa values across modalities (Text, Audio, and Video) for Part 2 on sexist categories.}
\label{tab:fleiss_kappa_part2_full}
\begin{adjustbox}{max width=\textwidth}
\begin{tabular}{lccccccccc}
\toprule
\textbf{Modality} & \textbf{Team} & \textbf{Sexist} & \textbf{Stereotype} & \textbf{Inequality} & \textbf{Discrimination} & \textbf{Objectification} & \textbf{Irony} & \textbf{Humor} \\
\midrule
\multirow{2}{*}{\textbf{Text}} 
 & Team 1 & 0.69 & 0.40 & 0.63 & 0.70 & 0.00 & 0.06 & 0.38 \\
 & Team 2 & 0.67 & 0.61 & 0.57 & 0.21 & 0.08 & 0.18 & 0.09 \\
\midrule
\multirow{2}{*}{\textbf{Audio}} 
 & Team 1 & 0.68 & 0.44 & 0.49 & 0.69 & 0.00 & 0.06 & 0.38 \\
 & Team 2 & 0.81 & 0.71 & 0.66 & 0.23 & 0.42 & 0.18 & 0.09 \\
\midrule
\multirow{2}{*}{\textbf{Video}} 
 & Team 1 & 0.81 & 0.50 & 0.72 & 0.73 & 0.49 & 0.06 & 0.38 \\
 & Team 2 & 0.67 & 0.50 & 0.45 & 0.49 & 0.30 & 0.18 & 0.09 \\
\bottomrule
\end{tabular}
\end{adjustbox}
\end{table}

\begin{table}[H]
\centering
\caption{Fleiss' Kappa values across modalities (Text, Audio, and Video) for Part 2 on non-sexist categories.}
\label{tab:fleiss_kappa_part2_meta}
\begin{adjustbox}{max width=\textwidth}
\begin{tabular}{lccccccc}
\toprule
\textbf{Modality} & \textbf{Team} & \textbf{Counter-speech} & \textbf{Reported sexism} & \textbf{Irony} & \textbf{Humor} \\
\midrule
\multirow{2}{*}{\textbf{Text}} 
 & Team 1 & 0.35 & 0.02 & 0.00 & 0.19 \\
 & Team 2 & 0.27 & 0.44 & -0.01 & 0.00 \\
\midrule
\multirow{2}{*}{\textbf{Audio}} 
 & Team 1 & 0.24 & 0.07 & 0.00 & 0.28 \\
 & Team 2 & 0.51 & 0.48 & 0.00 & 0.00 \\
\midrule
\multirow{2}{*}{\textbf{Video}} 
 & Team 1 & 0.38 & 0.09 & 0.00 & 0.11 \\
 & Team 2 & 0.39 & 0.34 & -0.01 & 0.00 \\
\bottomrule
\end{tabular}
\end{adjustbox}
\end{table}

\AppendixSection{Prompt Templates}\label{app:prompts}
The English translation of the video prompt is shown in Figure \ref{fig:prompt-video-en}. Figure~\ref{fig:prompt-txt-es} and Figure~\ref{fig:prompt-video-es} show the actual prompt templates used for the text and multimodal models in Spanish.

\begin{figure}[htb]
\caption{%
Prompt template for fine-grained sexism detection at the video level (English translation).
}

\begin{tcolorbox}[
    colback=textbox,
    colframe=headerbox,
    colbacktitle=headerbox,
    coltitle=white,
    title=Prompt for fine-grained sexism detection in video (EN),
    fontupper=\fontsize{8}{9}\selectfont,
    breakable,
    enhanced jigsaw,
    boxrule=0.4pt,
    arc=2pt,
    left=1mm,right=1mm,top=1mm,bottom=1mm,boxsep=0.5mm
]

\textcolor{cmint}{\textbf{Instruction:}}\newline
Analyze the content of this video and determine whether it is sexist. 
If the text of the video is sexist, the video must also be classified as sexist, even if the images do not appear offensive.
If it is sexist, indicate all categories from the following list that apply: ``stereotypes'', ``inequality'', ``discrimination'', ``objectification''. 
Some instances may belong to more than one category.\newline

\textcolor{cmint}{\textbf{Categories:}}\newline
\textbf{Stereotypes:} Descriptive traits allegedly distinguishing men and women; prescriptive traits enforcing societal gender roles.\newline
\textbf{Inequality:} Denial of past or present gender inequalities; opposition to feminism claiming it marginalizes men.\newline
\textbf{Discrimination:} Based on sexual orientation (e.g., homosexual, lesbian, intersex, bisexual, asexual, pansexual), gender identity (e.g., transgender), or against the LGBTQ+ community.\newline
\textbf{Objectification:} Portraying women as objects, evaluating their physical appearance, or criticizing them for not conforming to normative beauty standards.\newline

\textcolor{cmint}{\textbf{Response format:}}\newline
Answer only in JSON. If not sexist, return \texttt{"categories": []} with a brief justification.\newline

\textcolor{cmint}{\textbf{Video text:}}\newline

\textcolor{corange}{\texttt{[TEXT]}}

\medskip
\textit{Respond only with the JSON block. Do not include additional explanations.}

\end{tcolorbox}

\label{fig:prompt-video-en}
\end{figure}

\clearpage

\begin{figure}[htb]
\caption{Prompt template used for fine-grained sexism detection for text-only models.}
\begin{tcolorbox}[
    colback=textbox,
    colframe=headerbox,
    colbacktitle=headerbox,
    coltitle=white,
    title=Prompt template for fine-grained sexism detection (Text-only -- Spanish),
    fontupper=\fontsize{8}{9}\selectfont,
    breakable,
    enhanced jigsaw,
    boxrule=0.4pt,
    arc=2pt,
    left=1mm,right=1mm,top=1mm,bottom=1mm,boxsep=0.5mm
]
\textcolor{cmint}{\textbf{Instrucción:}}\newline
Determina si el texto es sexista. Si es sexista, indica todas las categorías que correspondan del siguiente listado. Algunas instancias pueden pertenecer a más de una categoría.\newline

\textcolor{cmint}{\textbf{Categorías:}}
\begin{itemize}\setlength{\itemsep}{1pt}\setlength{\parskip}{0pt}\setlength{\topsep}{2pt}
    \item \textbf{Estereotipos:}
    \begin{itemize}\itemsep0pt
        \item Formulación de propiedades descriptivas que supuestamente distinguen a hombres y mujeres basadas en estereotipos de género.
        \item Formulación de propiedades prescriptivas que hombres y mujeres deben cumplir para encajar en los roles de género establecidos por la sociedad.
    \end{itemize}
    \item \textbf{Desigualdad:}
    \begin{itemize}\itemsep0pt
        \item Contenido que niega la existencia de desigualdades (pasadas o presentes) entre hombres y mujeres.
        \item Contenido que se opone al feminismo, argumentando que este movimiento margina a los hombres.
    \end{itemize}
    \item \textbf{Discriminación:}
    \begin{itemize}\itemsep0pt
        \item Contenido que discrimina a personas por su orientación sexual (por ejemplo, personas homosexuales, lesbianas, intersexuales, bisexuales, asexuales y pansexuales), por su identidad de género, incluyendo críticas hacia quienes eligen un género distinto al asignado al nacer (por ejemplo, personas transgénero). También contenido que discrimina a personas LGBTQ+.
    \end{itemize}
    \item \textbf{Cosificación:}
    \begin{itemize}\itemsep0pt
        \item Contenido que presenta a las mujeres como objetos, evalúa su apariencia física o las critica por no ajustarse a estándares normativos de belleza.
    \end{itemize}
\end{itemize}

\textcolor{cmint}{\textbf{Formato de respuesta:}}\newline
Responde únicamente en formato JSON. Si el texto no es sexista, devuelve \texttt{"categorias": []} y una breve justificación.\newline

\textcolor{cmint}{\textbf{Ejemplo (sexista):}}
\begin{Verbatim}[fontsize=\footnotesize]
{
  "es_sexista": true,
  "categorias": ["estereotipo", "discriminación"],
  "justificacion": "El texto asigna roles de género 
tradicionales y discrimina a personas LGBTQ+."
}
\end{Verbatim}

\textcolor{cmint}{\textbf{Ejemplo (no sexista):}}
\begin{Verbatim}[fontsize=\footnotesize]
{
  "es_sexista": false,
  "categorias": [],
  "justificacion": "El texto no promueve estereotipos 
ni discriminación."
}
\end{Verbatim}

\textcolor{cmint}{\textbf{Texto a clasificar:}}\newline

\textcolor{corange}{\texttt{[TEXT]}}

\end{tcolorbox}
\label{fig:prompt-txt-es}
\end{figure}

\clearpage

\begin{figure}[htb]
\caption{Prompt template used for fine-grained sexism detection for multimodal language models.}
\begin{tcolorbox}[
    colback=textbox,
    colframe=headerbox,
    colbacktitle=headerbox,
    coltitle=white,
    title=Prompt for fine-grained sexism detection (Video -- Spanish),
    fontupper=\fontsize{8}{9}\selectfont,
    breakable,
    enhanced jigsaw,
    boxrule=0.4pt,
    arc=2pt,
    left=1mm,right=1mm,top=1mm,bottom=1mm,boxsep=0.5mm
]
\textcolor{cmint}{\textbf{Instrucción:}}\newline
Analiza el contenido del video y determina si es sexista. Si el texto del video es sexista, el video también debe clasificarse como sexista, incluso si las imágenes no parecen ofensivas. Si es sexista, indica todas las categorías aplicables. Algunas instancias pueden pertenecer a más de una categoría.\newline

\textcolor{cmint}{\textbf{Categorías:}}
\begin{itemize}[leftmargin=*,itemsep=1pt,topsep=2pt,parsep=0pt]
    \item \textbf{Estereotipos:}
    \begin{itemize}[itemsep=0pt,topsep=0pt]
        \item Formulación de propiedades descriptivas que supuestamente distinguen a hombres y mujeres basadas en estereotipos de género.
        \item Formulación de propiedades prescriptivas que hombres y mujeres deben cumplir para encajar en los roles de género establecidos por la sociedad.
    \end{itemize}
    \item \textbf{Desigualdad:}
    \begin{itemize}[itemsep=0pt,topsep=0pt]
        \item Contenido que niega la existencia de desigualdades (pasadas o presentes) entre hombres y mujeres.
        \item Contenido que se opone al feminismo, argumentando que este movimiento margina a los hombres.
    \end{itemize}
    \item \textbf{Discriminación:}
    \begin{itemize}[itemsep=0pt,topsep=0pt]
        \item Contenido que discrimina a personas por su orientación sexual (por ejemplo, personas homosexuales, lesbianas, intersexuales, bisexuales, asexuales y pansexuales), por su identidad de género, incluyendo críticas hacia quienes eligen un género distinto al asignado al nacer (por ejemplo, personas transgénero). También contenido que discrimina a personas LGBTQ+.
    \end{itemize}
    \item \textbf{Cosificación:}
    \begin{itemize}[itemsep=0pt,topsep=0pt]
        \item Contenido que presenta a las mujeres como objetos, evalúa su apariencia física o las critica por no ajustarse a estándares normativos de belleza.
    \end{itemize}
\end{itemize}

\textcolor{cmint}{\textbf{Formato de respuesta:}}\newline
Responde solo en JSON. Si no es sexista, devuelve \texttt{"categorias": []} y una breve justificación.\newline

\textcolor{cmint}{\textbf{Ejemplo (sexista):}}
\begin{Verbatim}[fontsize=\footnotesize]
{
  "es_sexista": true,
  "categorias": ["estereotipo", "discriminación"],
  "justificacion": "El video asigna roles de género 
tradicionales y discrimina a personas LGBTQ+."
}
\end{Verbatim}

\textcolor{cmint}{\textbf{Ejemplo (no sexista):}}
\begin{Verbatim}[fontsize=\footnotesize]
{
  "es_sexista": false,
  "categorias": [],
  "justificacion": "El video no promueve estereotipos 
ni discriminación."
}
\end{Verbatim}

\textcolor{cmint}{\textbf{Texto del video:}}\newline

\textcolor{corange}{[TEXT]}

\medskip

\textit{Responde solo con el bloque JSON.}

\end{tcolorbox}
\label{fig:prompt-video-es}
\end{figure}

\AppendixSection{Results}\label{app:results}

Table~\ref{tab:binary_appendix} and Table~\ref{tab:fg_full} include the Format Error Rate (FoER) and, for the binary task, the Valid Accuracy. Accuracy (All) as reported inTable \ref{tab:results_binary}, is computed over every instance, counting invalid or missing outputs as incorrect. Valid Accuracy, by contrast, is computed only on the subset of valid model outputs, showing performance when the model produces a well-formed response. As noted in the main text, FoER was negligible ($<=0.07$) for nearly all models. The only exception was Qwen-32B, which produced malformed outputs in up to 35\% of cases on the binary task (P2). Table \ref{tab:fg_perclass} shows the per-class F1 scores for fine-grained sexism categories.

\begin{table}[H]
\centering
\caption{Binary sexism detection results (full metrics). 
All = Accuracy (All), Val = Accuracy (Valid), FoER = Format Error Rate. 
Results are shown for Part 1 (P1, MuSeD: TikTok \& BitChute) and Part 2 (P2, YouTube Shorts).}
\small
\begin{tabular}{lcccccc}
\toprule
& \multicolumn{3}{c}{\textbf{P1}} & \multicolumn{3}{c}{\textbf{P2}} \\
\cmidrule(lr){2-4} \cmidrule(lr){5-7}
\textbf{Model} & All & Val & FoER & All & Val & FoER \\
\midrule
\multicolumn{7}{l}{\textit{Smaller LLMs}} \\
Llama-3-8B-Instruct & 61.75 & 63.16 & 0.05 & 67.00 & 67.00 & 0.07 \\
Qwen2.5-3B-Instruct & 61.00 & 61.00 & 0.00 & 66.59 & 66.59 & 0.00 \\
Salamandra-7B-Instruct & 55.00 & 60.05 & 0.00 & 55.00 & 60.05 & 0.00 \\
\midrule
\multicolumn{7}{l}{\textit{Larger LLMs}} \\
Llama-3-70B-Instruct & 83.25 & 84.36 & 0.02 & 75.06 & 75.06 & 0.05 \\
Qwen2.5-32B-Instruct & 72.25 & 78.42 & 0.18 & 73.84 & 73.84 & 0.35 \\
GPT-4o & 78.00 & 78.20 & 0.00 & 76.87 & 76.87 & 0.00 \\
Claude-3.7-Sonnet & 85.00 & 85.00 & 0.00 & 80.84 & 80.84 & 0.00 \\
Gemini-2.0-Flash & 72.50 & 72.43 & 0.00 & 72.43 & 72.43 & 0.00 \\
\midrule
\multicolumn{7}{l}{\textit{Multimodal LLMs}} \\
GPT-4o (V+L) & 79.50 & 79.29 & 0.01 & 78.12 & 78.12 & 0.01 \\
Claude-3.7-Sonnet (V+L) & 85.75 & 85.96 & 0.00 & 83.64 & 83.64 & 0.00 \\
Gemini-2.0-Flash (V+L) & 71.00 & 71.18 & 0.00 & 71.29 & 71.29 & 0.01 \\
Gemini-2.0-Flash (Video) & 68.25 & 68.42 & 0.00 & 72.66 & 72.66 & 0.00 \\
\bottomrule
\end{tabular}
\label{tab:binary_appendix}
\end{table}

\begin{table}[H]
\centering
\caption{Format Error Rates (FoER) for fine-grained sexism detection. Values correspond to Part 1 (P1: TikTok \& BitChute) and Part 2 (P2: YouTube Shorts).}
\small
\begin{tabular}{lcc}
\toprule
\textbf{Model} & \textbf{P1 FoER} & \textbf{P2 FoER} \\
\midrule
\multicolumn{3}{l}{\textit{Smaller LLMs}} \\
Llama-3-8B-Instruct       & 0.07 & 0.11 \\
Qwen2.5-3B-Instruct       & 0.00   & 0.00   \\
Salamandra-7B-Instruct    & 0.00   & 0.00   \\
\midrule
\multicolumn{3}{l}{\textit{Larger LLMs}} \\
Llama-3-70B-Instruct      & 0.03 & 0.06 \\
Qwen2.5-32B-Instruct      & 0.21 & 0.37 \\
GPT-4o                    & 0.01 & 0.01 \\
Claude-3.7-Sonnet         & 0.00 & 0.00 \\
Gemini-2.0-Flash          & 0.00 & 0.00 \\
\midrule
\multicolumn{3}{l}{\textit{Multimodal LLMs}} \\
GPT-4o (V+L)              & 0.00 & 0.00 \\
Claude-3.7-Sonnet (V+L)   & 0.01 & 0.00 \\
Gemini-2.0-Flash (Video)  & 0.01 & 0.00 \\
Gemini-2.0-Flash (V+L)    & 0.01 & 0.00 \\
\bottomrule
\end{tabular}
\label{tab:fg_full}
\end{table}


\begin{table}[H]
\centering
\small
\caption{Per-class F1 scores for fine-grained sexism categories. 
P1 = Part 1 (TikTok \& BitChute), P2 = Part 2 (YouTube Shorts). 
St = Stereotypes, In = Inequality, Di = Discrimination, Ob = Objectification. Best scores are in \textbf{bold}.}
\begin{tabular}{lcccccccc}
\toprule
 & \multicolumn{4}{c}{\textbf{P1: TikTok \& BitChute}} & 
   \multicolumn{4}{c}{\textbf{P2: YouTube Shorts}} \\
\cmidrule(lr){2-5} \cmidrule(lr){6-9}
\textbf{Model} & St & In & Di & Ob & St & In & Di & Ob \\
\midrule
Random & 27.25 & 19.00 & 7.50 & 1.75 & 25.93 & 26.87 & 14.02 & 2.80 \\
\midrule
\multicolumn{9}{l}{\textit{Smaller LLMs}} \\
Llama-3-8B-Instruct & 73.09 & 61.20 & 38.30 & 17.91 & 58.44 & 66.06 & 24.00 & 7.84 \\
Qwen2.5-3B-Instruct & 72.45   & 5.00   & 19.89   & 3.13   & 64.36   & 1.72   & 34.08   & 8.70 \\
Salamandra-7B-Instruct & 72.30 & 55.65   & 40.00   & 13.64   & 61.83   & 68.87   & 45.71   & 11.76 \\
\midrule
\multicolumn{9}{l}{\textit{Larger LLMs}} \\
Llama-3-70B-Instruct & 70.29 & 44.44 & 33.57 & 11.11 & 58.29 & 40.00 & 39.51 & 26.67 \\
Qwen2.5-32B-Instruct & 71.96 & 63.09 & 43.33 & 20.51 & 50.53 & 42.86 & 38.55 & 14.81 \\
GPT-4o & 82.93 & 75.16 & 77.61 & 12.77 & 64.86 & 75.25 & 79.65 & 33.33 \\
Claude-3.7-Sonnet & 67.04 & 40.00 & 64.15 & \textbf{35.29} & 61.36 & 42.38 & 66.67 & 38.10 \\
Gemini-2.0-Flash & 45.95 & 63.95 & 68.97 & 25.00 & 31.43 & 45.00 & 60.87 & \textbf{56.00} \\
\midrule
\multicolumn{9}{l}{\textit{Multimodal LLMs}} \\
GPT-4o (V+L) & \textbf{89.30} & \textbf{75.56} & 76.67 & 20.00 & 69.79 & 75.65 & 74.29 & 50.00 \\
Claude-3.7-Sonnet (V+L) & 88.70 & 74.29 & \textbf{82.35} & 25.00 & \textbf{77.78} & \textbf{82.67} & \textbf{81.08} & 50.00 \\
Gemini-2.0-Flash (V+L) & 84.06 & 72.15 & 60.71 & 18.18 & 69.23 & 67.02 & 72.16 & 50.00 \\
Gemini-2.0-Flash (Video) & 79.65 & 65.81 & 43.64 & 0.00 & 70.54 & 53.63 & 62.00 & 33.33 \\
\bottomrule
\end{tabular}
\label{tab:fg_perclass}
\end{table}


\begin{table*}[t]
\centering
\small
\caption{Sample of comparison between the generated explanations of Claude-3.7-Sonnet (V+L) and human evaluators. Ratings are based on a 5-point Likert scale. The scores refer to three categories: \textbf{R} for Relevance, \textbf{I} for Informativeness, \textbf{G} for Groundness. Each annotator received the binary and fine-grained labels of the sexist types for each video.}
\resizebox{\textwidth}{!}{
\begin{tabular}{p{5cm} p{5cm} c c c c c c}
\toprule
\textbf{Explanation 1} & \textbf{Explanation 2} &
\textbf{R1} & \textbf{I1} & \textbf{G1} &
\textbf{R2} & \textbf{I2} & \textbf{G2} \\
\midrule

El contenido presenta de forma irónica supuestos 'privilegios masculinos' que son en realidad desventajas, minimizando las desigualdades estructurales que enfrentan las mujeres y reforzando estereotipos de género sobre roles laborales y responsabilidades familiares. 
\newline
\textit{The content ironically presents supposed 'male privileges' which are actually disadvantages, minimizing structural inequalities faced by women and reinforcing gender stereotypes about work roles and family responsibilities.} &
El video presenta contenido sexista puesto que pone en duda el privilegio masculino, argumentando como ocupan puesto de mayor riesgo físico, además de que se tienen que jubilar más tarde, acostumbran a perder los juicios por potestas de los hijos, etc.
\newline
\textit{The video presents sexist content as it questions male privilege, arguing that men occupy higher-risk positions, retire later, and often lose child custody cases.} &
5 & 5 & 4 & 3 & 3 & 2 \\

\midrule

El video explica conceptos relacionados con orientación sexual y critica la heteronorma (expectativa de que las relaciones sean heterosexuales). Promueve el entendimiento de la diversidad sexual y reconoce diferentes realidades, sin perpetuar estereotipos sexistas.
\newline
\textit{The video explains concepts related to sexual orientation and critiques heteronormativity (the expectation that relationships are heterosexual). It promotes understanding of sexual diversity and acknowledges different realities without perpetuating sexist stereotypes.} &
Este vídeo no contiene contenido sexista, ya que hace divulgación sobre la orientación sexual. No muestra estereotipos ni discriminación por la identidad sexual. Explica de forma divulgativa conceptos que usa, como el de cisgénero, y muestra que hay muchas maneras de entender el deseo y la orientación sexual.
\newline
\textit{This video does not contain sexist content, as it provides educational information about sexual orientation. It does not show stereotypes or discrimination based on sexual identity. It explains, in an educational way, concepts it uses—such as "cisgender" and demonstrates that there are many ways to understand desire and sexual orientation.} &
5 & 3 & 4 & 5 & 5 & 5 \\

\bottomrule
\end{tabular}
}
\label{tab:comparison_explanations}
\end{table*}

\begin{table*}[t]
\centering
\small
\caption{Wilcoxon signed-rank test results for human vs. model explanations. The metrics are: \textbf{R} for Relevance, \textbf{I} for Informativeness, and \textbf{G} for Groundness. Values with a statistically significant difference are shown in bold.}
\small
{%
\begin{tabular}{l l r r r r r r}
\hline
\textbf{\shortstack{Evaluator}} & \textbf{\shortstack{Metric}} & \textbf{\shortstack{W}} & \textbf{\shortstack{p-value}} & \textbf{\shortstack{Average\\Human}} & \textbf{\shortstack{Average\\Model}} & \textbf{\shortstack{Median\\Human}} & \textbf{\shortstack{Median\\Model}} \\
\hline
E1 & R & 97.5  & 0.12 & 4.03 & 4.46 & 4.0 & 5.0 \\
E2 & R & 8.0   & 0.26 & 4.34 & 4.43 & 4.0 & 5.0 \\
E3 & R & 30.0  & \textbf{0.03} & 4.66 & 4.37 & 5.0 & 5.0 \\
\hline
E1 & I & 112.0 & 0.06 & 3.94 & 3.40 & 4.0 & 3.0 \\
E2 & I & 133.5 & 0.40 & 3.89 & 4.03 & 4.0 & 4.0 \\
E3 & I & 127.0 & 0.32 & 4.09 & 3.89 & 4.0 & 4.0 \\
\hline
E1 & G & 52.5  & 1.00 & 4.34 & 4.34 & 5.0 & 5.0 \\
E2 & G & 126.5 & 1.00 & 4.40 & 4.40 & 4.0 & 4.0 \\
E3 & G & 39.0  & 0.11 & 4.17 & 3.91 & 4.0 & 4.0 \\
\hline
\end{tabular}%
}
\label{tab:wilcoxon_results}
\end{table*}




\clearpage
\bibliographystyle{compling}

\bibliography{COLI_template}

\end{document}